\documentclass[runningheads]{llncs}

\usepackage[final]{eccv}

\usepackage{eccvabbrv}

\usepackage{graphicx}
\usepackage{booktabs}
\usepackage{multirow}
\usepackage{amsmath}
\usepackage{amssymb}
\usepackage{xcolor}
\usepackage{pifont}
\usepackage{tikz}
\usetikzlibrary{positioning,decorations.pathreplacing,calligraphy}
\usepackage[accsupp]{axessibility}  

\usepackage{hyperref}
\usepackage{orcidlink}

\newcommand{\yes}{\textcolor{teal}{\ding{51}}}
\newcommand{\no}{\textcolor{red!70!black}{\ding{55}}}

\definecolor{findingframe}{RGB}{38,148,132}
\definecolor{findingback}{RGB}{233,247,244}
\newsavebox{\findingbox}
\newenvironment{finding}
  {\par\smallskip\noindent\begin{lrbox}{\findingbox}%
   \begin{minipage}{\dimexpr\linewidth-2\fboxsep-2\fboxrule\relax}\small}
  {\end{minipage}\end{lrbox}%
   \setlength{\fboxrule}{0.9pt}%
   \noindent\fcolorbox{findingframe}{findingback}{\usebox{\findingbox}}\par\smallskip}

\newcommand{\IST}{\textsc{IST}}
\newcommand{\STI}{\textsc{STI}}
\newcommand{\SIT}{\textsc{SIT}}
\newcommand{\STIT}{\textsc{STIT}}
\newcommand{\SITIT}{\textsc{SITIT}}
\newcommand{\SITITr}{\textsc{SITIT-r}}
\newcommand{\SITITp}{\textsc{SITIT-p}}

\newsavebox{\imgiconbox}
\newsavebox{\qiconbox}
\newcommand{\imgicon}{\usebox{\imgiconbox}}
\newcommand{\qicon}{\usebox{\qiconbox}}

\begin{document}

\sbox{\imgiconbox}{\tikz[baseline=-0.55ex,line width=0.3pt]{%
  \begin{scope}
    \clip[rounded corners=0.16ex] (0,-0.05ex) rectangle (1.85ex,1.35ex);
    \fill[blue!20] (0,-0.05ex) rectangle (1.85ex,1.35ex);
    \fill[yellow!85!orange] (1.42ex,1.02ex) circle (0.19ex);
    \fill[green!45!teal] (-0.1ex,-0.05ex) -- (0.52ex,0.66ex) -- (0.86ex,0.34ex)
      -- (1.34ex,0.84ex) -- (1.9ex,0.46ex) -- (1.9ex,-0.05ex) -- cycle;
  \end{scope}
  \draw[rounded corners=0.16ex, black!55] (0,-0.05ex) rectangle (1.85ex,1.35ex);}}
\sbox{\qiconbox}{\tikz[baseline=-0.55ex,line width=0.3pt]{%
  \fill[orange!90!red,rounded corners=0.34ex] (0,0.3ex) rectangle (1.45ex,1.46ex);
  \fill[orange!90!red] (0.36ex,0.42ex) -- (0.24ex,-0.02ex) -- (0.7ex,0.42ex) -- cycle;
  \node[inner sep=0pt,font=\tiny\bfseries,text=white] at (0.72ex,0.9ex) {?};}}

\title{Ask Twice, Look Twice: Prompt Echoing Resolves the Question-First Paradox in Vision-Language Models}
\titlerunning{Ask Twice, Look Twice: Echoing in VLMs}


\author{Rakshanda Hassan Abhinandan \and John Galeotti \and Deva Ramanan \and \\Gautam Rajendrakumar Gare}

\authorrunning{Abhinandan et. al.}

\institute{Carnegie Mellon University, USA}

\maketitle

\begin{abstract}
Where should the question go in a vision-language model (VLM) prompt: before the image or after it? Intuition says before: knowing what is asked should tell the model where to look. Yet across visual question answering benchmarks, question-first prompting consistently underperforms the image-first ordering recommended for frontier VLMs, a phenomenon we term the \emph{question-first paradox}. We trace the paradox to a conflict between two stages of VLM computation. Logit-lens and attention probes show the intuition is half right: a question placed before the image genuinely steers perception, moving image patch representations toward question-relevant concepts. The failure lies downstream. Stranded behind hundreds of image tokens, the question is barely attended by the answer token, which instead commits to image-driven (often wrong) answers; a causal attention knockout confirms that the answer reads the question only when the question follows the image. Better perception is squandered by worse access. The diagnosis yields a training-free fix: \emph{question echoing}, restating the question on both sides of the image so that one copy steers perception while the other is read out at answer time. A similar division of labor appears in a fifty-year-old finding on human ``adjunct questions'', where repeating a question before and after a passage aids comprehension more than either position alone. \emph{Echoing the image} as well brings further gains, restoring the whole-image view a causal decoder otherwise loses. The paradox holds across five open VLMs, costing up to 17.5 group-accuracy points. Echoed prompts recover most of the gap and, on the harder benchmarks (NaturalBench and Winoground), surpass the best single-pass ordering by up to 19 group-accuracy points on Winoground, with no training, fine-tuning, or architecture change. The paradox exposes a tension in multimodal prompting between steering what a model \emph{sees} and preserving access to what it was \emph{asked}; echoing resolves it with prompt design alone.
Our code and results are available on our project page: \url{https://rakshanda-cmu.github.io/ask-twice-look-twice/}.
\end{abstract}

\section{Introduction}
\label{sec:intro}

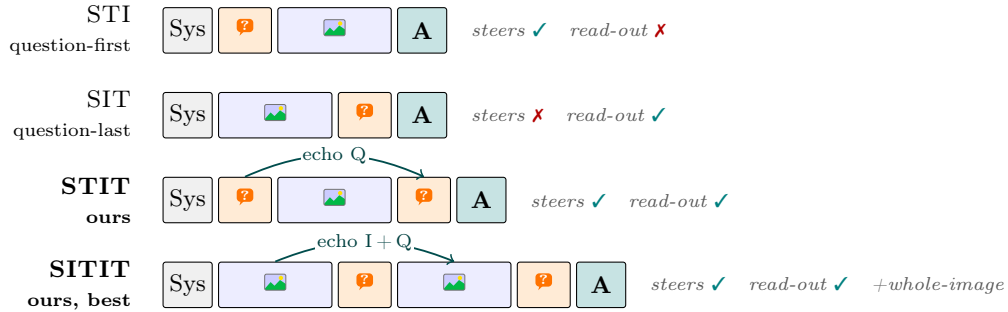
\begin{figure}[t]
  \centering
  \begin{tikzpicture}[
    font=\footnotesize,
    blk/.style={draw, minimum height=6mm, rounded corners=1pt, align=center, inner sep=1.5pt},
    sys/.style={blk, fill=gray!12, minimum width=6.5mm},
    task/.style={blk, fill=orange!16, minimum width=7mm, font=\footnotesize\bfseries},
    img/.style={blk, fill=blue!7, minimum width=15mm},
    ans/.style={blk, fill=teal!22, minimum width=6.5mm, font=\footnotesize\bfseries},
    rowlab/.style={anchor=east, align=right},
    note/.style={anchor=west, font=\scriptsize\itshape, text=black!65},
    echo/.style={->, teal!60!black, line width=0.7pt},
    echolab/.style={midway, fill=white, inner sep=0.7pt, font=\scriptsize,
                    text=teal!50!black},
  ]
    \def\ya{0}\def\yb{-1.12}\def\yc{-2.24}\def\yd{-3.36}

    \node[rowlab] at (-0.3,\ya) {\textsc{STI}\\\scriptsize question-first};
    \node[sys,anchor=west] (as) at (0,\ya) {Sys};
    \node[task,right=0.7mm of as] (aq) {\qicon};
    \node[img,right=0.7mm of aq] (ai) {\imgicon};
    \node[ans,right=0.7mm of ai] (aa) {A};
    \node[note,right=2mm of aa.east] {steers \yes\quad read-out \no};

    \node[rowlab] at (-0.3,\yb) {\textsc{SIT}\\\scriptsize question-last};
    \node[sys,anchor=west] (bs) at (0,\yb) {Sys};
    \node[img,right=0.7mm of bs] (bi) {\imgicon};
    \node[task,right=0.7mm of bi] (bq) {\qicon};
    \node[ans,right=0.7mm of bq] (ba) {A};
    \node[note,right=2mm of ba.east] {steers \no\quad read-out \yes};

    \node[rowlab] at (-0.3,\yc) {\textbf{\textsc{STIT}}\\\scriptsize \textbf{ours}};
    \node[sys,anchor=west] (cs) at (0,\yc) {Sys};
    \node[task,right=0.7mm of cs] (cq) {\qicon};
    \node[img,right=0.7mm of cq] (ci) {\imgicon};
    \node[task,right=0.7mm of ci] (cq2) {\qicon};
    \node[ans,right=0.7mm of cq2] (ca) {A};
    \node[note,right=2mm of ca.east] {steers \yes\quad read-out \yes};
    \draw[echo] (cq.north) to[bend left=24] node[echolab]{echo Q} (cq2.north);

    \node[rowlab] at (-0.3,\yd) {\textbf{\textsc{SITIT}}\\\scriptsize \textbf{ours, best}};
    \node[sys,anchor=west] (ds) at (0,\yd) {Sys};
    \node[img,right=0.7mm of ds] (di) {\imgicon};
    \node[task,right=0.7mm of di] (dq) {\qicon};
    \node[img,right=0.7mm of dq] (di2) {\imgicon};
    \node[task,right=0.7mm of di2] (dq2) {\qicon};
    \node[ans,right=0.7mm of dq2] (da) {A};
    \node[note,right=2mm of da.east] {steers \yes\quad read-out \yes\quad +whole-image};
    \draw[echo] (di.north) to[bend left=22] node[echolab]{echo I\,+\,Q} (di2.north);
  \end{tikzpicture}
  \caption{\textbf{Question placement, not question presence, decides the
  answer.} A VLM reads system (\textbf{Sys}), image (\imgicon), and
  question (\qicon) tokens as one stream and generates from the final
  answer position (\textbf{A}). Each row is marked on the two mechanisms a
  prompt must serve: whether the question \emph{steers} what the image encodes,
  and whether it is \emph{read out} at the answer. Question-first (\STI{})
  steers but is not read out: it strands the question before a long image span,
  so the answer barely attends to it and commits early to an image-anchored,
  often-wrong response. Question-last (\SIT{}) is read out but does not steer,
  recovering most of the loss. Training-free \emph{question echoing} (\STIT{}) restates the
  question on both sides of the image, keeping the steering while
  restoring question--answer adjacency; re-presenting the image with it
  (\SITIT{}) also grants the causal decoder a whole-image view and gives
  the best accuracy.}
  \label{fig:teaser}
\end{figure}

Every VLM prompt makes a silent decision: the question either precedes or
follows the image. Modern vision-language models consume a single interleaved
sequence of system, image, and task tokens, so the ordering is a free
design choice, and today it is decided by folklore.
Most model cards recommend image-first~\cite{AprilGemini,ImageLearn,QwenVL3Card,QwenVL25Card,Gemma3Card,InternVL3Card,LLaVA15Card}, while other prompting practices vary.
A natural hypothesis, borrowed from how people
inspect a scene \cite{Buswell1935HowArt}, says the folklore has it
backwards: the question should come first, because a model that knows
what is asked can decide where to look and what to encode.

The hypothesis fails, and it fails in an instructive way. Consider the
two orderings that hold content fixed and move only the question: \STI{}
(System, Task, Image) places the question before the image, \SIT{}
(System, Image, Task) after. On NaturalBench with Qwen3-VL-8B,
question-first scores 8.1 group-accuracy points below question-last
(0.270 vs.\ 0.351). The same gap appears on POPE and Winoground and on
three further VLM families (Qwen2.5-VL, InternVL3, LLaVA), reaching 17.5
points; it is muted only on Gemma-3-27B. We call this the
\emph{question-first paradox}: the ordering that should inform perception
is the one that hurts (\cref{fig:teaser}).

This paper opens the model to explain the paradox, then turns the
explanation into a fix. The resolution is a dissociation between two
stages of computation that prompt design conflates. Using the logit
lens~\cite{2020InterpretingLessWrong} and layer-wise probes, we show
that question-first prompting \emph{does} steer perception: image patches
decode to more question-relevant concepts, and the answer position
attends more strongly to the image. The steering is real, and it is
wasted. Stranded before hundreds of image tokens, the question is
under-read by the decoder, and the answer position commits early to an
image-anchored, often wrong token before a question representation
reaches it. \textbf{Better perception is squandered by worse access.}

Once stated this way, the fix writes itself. \emph{Question echoing}
(\STIT{}) restates the question once before and once after the image: the
first copy steers, the second restores question--answer adjacency.
\emph{Image echoing} (\SITIT{}) re-presents the image alongside the
repeated question; under a causal mask every patch of the second copy
attends over the entire first copy, recovering the bidirectional
whole-image read that the vision encoder computes but the causal decoder
discards. Both are pure prompt edits: no fine-tuning, no decoding change,
no architecture change.

Strikingly, the fix has a fifty-year-old precedent in human learners. In educational
psychology's \emph{adjunct question} literature, repeating the same
question before and after a prose passage aids comprehension more than
either position alone: the prequestion directs attention, the
postquestion consolidates at test
time~\cite{Boyd1974RepeatedLearning,Hamaker1986TheLearning}. This is the
division of labor our probes uncover inside the model, suggesting the
two-mechanism structure of echoing is a general property of sequential
comprehension rather than an artifact of one architecture.

\begin{figure}[t]
  \centering
  {\footnotesize Q: \emph{``Is the adult chasing a child?''} \quad (ground truth: \textbf{Yes})}\\[3pt]
  \includegraphics[width=\linewidth]{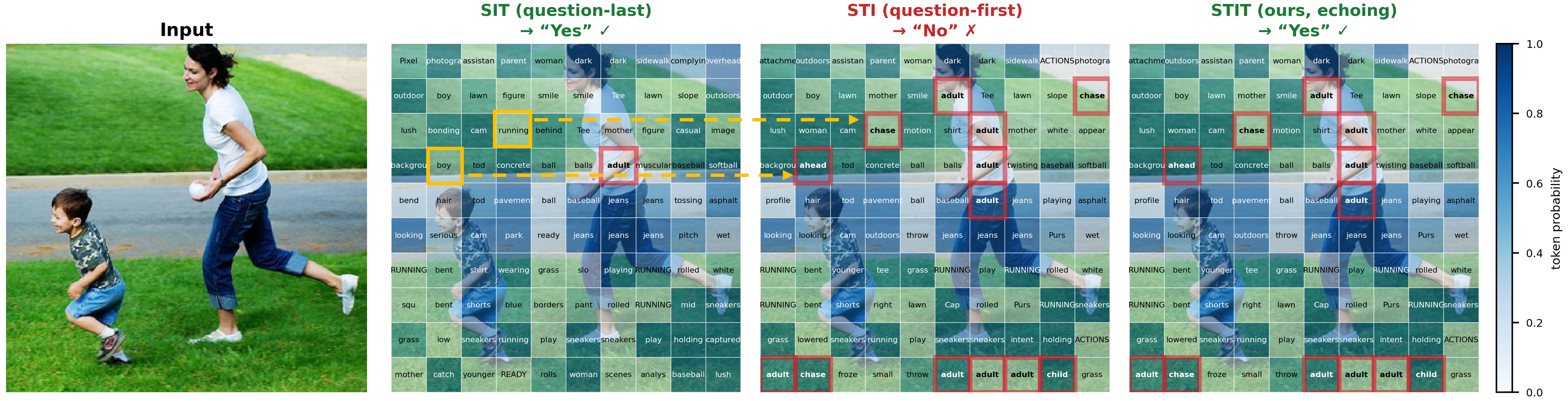}
  \caption{\textbf{Steering happens, yet is not read out.} Each cell shows the
  vocabulary token its image patch's final-layer (layer 32) hidden state decodes
  to (logit lens, Qwen3-VL), coloured by its probability (blue scale); red
  \textbf{rings} mark patches whose token reflects the question (\emph{adult},
  \emph{chase}, \emph{child}, \emph{ahead}). Question-last (\SIT{}) leaves the
  patches generic and answers correctly; question-first (\STI{}) steers the
  \emph{same} patches toward the question, turning \emph{boy} and \emph{running}
  into \emph{ahead} and \emph{chase}, yet answers wrong: steering happens, but is
  not read out. Our fix (echoing, \STIT{}; \cref{sec:method}) keeps the steering
  and answers correctly.}
  \label{fig:steering}
\end{figure}

\noindent\textbf{Contributions.} We make four contributions. First, we
establish the \emph{question-first paradox}: with content held fixed, placing
the question before the image degrades accuracy across three benchmarks and
five open VLMs, against the intuition that it should help (\cref{sec:paradox}).
Second, we diagnose it, showing that question-first prompting genuinely steers
perception yet is not read out, and confirming with a causal attention
knockout that the answer is computed from the question only when the question
follows the image; perceptual steering and decoder access are dissociable, and
the paradox lives entirely in access (\cref{sec:analysis}). Third, we turn the
diagnosis into a training-free fix, \emph{question echoing}, which restates the
question after the image to restore read-out while preserving the steering,
together with an \emph{image-echoing} variant that recovers a whole-image read
under the causal mask (\cref{sec:method,sec:bidir}). Finally, we ablate the
design, ruling out ``more image tokens'' and ``more pre-answer compute'' and
isolating question adjacency and in-distribution image-first pairing as the
active ingredients (\cref{sec:ablation}).

The practical message: one duplicated question line recovers most of the
question-first gap at no real cost, and a duplicated (image, question) pair
exceeds the best single-pass ordering. The scientific message: visual steering and
answer read-out are distinct mechanisms with different positional
preferences, and prompt design must serve both. Prompt ordering can be a
measured, mechanistically grounded science rather than folklore; this
paper is a case study in making it one. Our code and a project website are
included in the supplementary material.

\section{Related Work}
\label{sec:related}

\paragraph{Prompt ordering in VLMs.}
Instruction-tuned VLMs such as LLaVA~\cite{Liu2023VisualTuningb},
Qwen2.5-VL~\cite{Bai2025Qwen2.5-VLReportb}, and Gemma~3~\cite{Team2025GemmaReport} flatten
image and text into a single token stream consumed by a causal decoder,
so the relative placement of visual tokens and the question is an
unavoidable design choice. It is rarely treated as a variable in its own
right: order-sensitivity studies focus on in-context
demonstrations~\cite{Tan2024OrderModels} or report image-first versus
text-first accuracy as an unexplained
ablation~\cite{Ismithdeen2025Promptception:Prompts}, and model cards recommend
image-first~\cite{QwenVL3Card,QwenVL25Card,Gemma3Card,InternVL3Card,LLaVA15Card}
without saying why; no prior work establishes \emph{why} question-first hurts. Concurrent work by Gupta and Gandelsman~\cite{Gupta2026TestTimeModality}
independently documents the same failure and localizes it with activation
patching to a narrow mid-network region, then closes it with asymmetric
test-time training. We treat prompt ordering as a first-class object and differ
in two ways: we localize a mechanism rather than a region, causally dissociating
intact perceptual steering (\cref{sec:steering}) from a failed read-out
(\cref{sec:causal}), and our fix is training-free (question echoing,
\cref{sec:method}).

\paragraph{Question repetition and re-reading.}
In text-only LLMs, re-presenting the query has a mixed record: re-reading
prompts improve reasoning~\cite{Xu2023Re-ReadingModels}, repeated question
tokens draw more attention~\cite{HanREADREADING}, and prompt repetition helps
most when the query sits far from the answer
position~\cite{Leviathan2025PromptLLMs}, yet controlled studies also find
the gains insignificant~\cite{Shaier2024AskingQuestionsb}. These works neither involve
vision nor explain when repetition should help. In VLMs the long image
span makes the query--answer distance structural rather than incidental;
we identify decoder under-reading of the distant question as the
mechanism and show echoing yields large, consistent gains exactly where
the text-only literature is equivocal. Image echoing has no text-only
analogue. The failure we document is related to ``lost in the
middle''~\cite{Liu2023LostContexts}, where decoder-only models under-use
context far from the query, but with a twist the text-only setting
lacks: the distant question still steers perception successfully, and
the failure arises only at answer generation.

\paragraph{Interpreting VLM representations.}
The logit lens~\cite{2020InterpretingLessWrong} projects intermediate
hidden states through the output embedding to read what each layer
represents; applied to VLMs, it shows that visual tokens progressively
align with interpretable vocabulary~\cite{Neo2024TowardsModels}. We build on
this tool but ask a different question: not what visual tokens encode, but
how prompt ordering changes what they encode and whether the decoder reads
it out.

\section{The Question-First Paradox}
\label{sec:paradox}

\subsection{Notation and setup}
A prompt has three sections: the system message (S), the image (I,
expanded into many visual tokens), and the task or question (T). We write
an ordering as the sequence of section letters in token order; generation
begins after the last letter. The system message is a fixed prefix
prepended by default (\cref{sec:supp-sysfirst} justifies this choice).
The two single-question orderings hold content fixed and move only the
question:
\begin{align}
\STI{}\;&=\;\text{System},\ \text{Task},\ \text{Image} && \text{(question-first)},\\
\SIT{}\;&=\;\text{System},\ \text{Image},\ \text{Task} && \text{(question-last)}.
\end{align}
They are anagrams. \STI{} separates the question from the answer position
by the entire image span; \SIT{} places it immediately before the answer.

All orderings are produced by one order-aware input builder shared across
models, so variants differ only in the arrangement of identical content.
We evaluate on NaturalBench~\cite{Li2024NaturalBench:Samples},
POPE~\cite{Li2023EvaluatingModels}, and
Winoground~\cite{Thrush2022Winoground:Compositionality} (\cref{sec:related}),
with greedy decoding, a 16-token answer budget, and a fixed system prompt
(\cref{sec:supp-setup}). Qwen3-VL-8B (primary, with all ablations) and
Gemma-3-27B carry the full analysis; three further families (Qwen2.5-VL,
InternVL3, LLaVA-1.5) test generality. We headline the most demanding
metric, group accuracy on NaturalBench and Winoground, which a single wrong
pair drives to zero and on which language priors do not help. For adjacent
orderings we report paired significance over per-group correctness
(two-sided exact McNemar; paired bootstrap 95\% CI, 5{,}000 resamples).

\subsection{The phenomenon}
\label{sec:phenomenon}
\Cref{tab:paradox} isolates the paradox: hold the content fixed, move only
the question, and question-first loses. On our primary model the gap is
8.1 group-accuracy points on NaturalBench and larger still on the
compositional Winoground, all highly significant and free, since the token
count is identical (per-comparison tests in \cref{sec:supp-results}). The effect is not a quirk of one model. It reproduces on
four of five VLMs and on all three benchmarks, and it is largest on the
weaker models, where question-first can collapse a model to a single constant
answer (\cref{tab:paradox}). The lone exception is Gemma-3-27B, where the two
orderings are statistically indistinguishable; even there the remedy of
\cref{sec:exp} still helps, so what generalizes is not the deficit but its
cure.

\begin{table}[tb]
  \caption{\textbf{The question-first paradox, across architectures.} Identical
  content; only the question moves. NaturalBench group accuracy for
  question-first (\STI{}) vs.\ question-last (\SIT{}), and the paradox gap
  $\Delta=\SIT{}-\STI{}$ on each benchmark (positive $\Delta$ means question-first
  is worse); \textbf{bold} marks each model's largest gap. Paired significance for
  every gap is in \cref{tab:significance}. The paradox is large and consistent on
  four models and muted only on Gemma. All three benchmarks show the same paradox
  on the added models.
  It is most dramatic on LLaVA-1.5, where question-first collapses to a constant
  ``Yes'' on every benchmark (NaturalBench group 0.000, POPE accuracy 0.500,
  Winoground group 0.000) and question-last repairs it (POPE 0.871), the largest
  POPE gap of any model ($+0.371$). The Qwen2.5-VL and InternVL3 gaps (POPE up to
  $+0.057$, Winoground up to $+0.210$) also exceed the two primary models.}
  \label{tab:paradox}
  \centering
  \small
  \begin{tabular}{@{}lccccc@{}}
    \toprule
    & \multicolumn{2}{c}{NatBench acc} & \multicolumn{3}{c}{paradox gap $\Delta=\SIT{}-\STI{}$}\\
    \cmidrule(lr){2-3}\cmidrule(lr){4-6}
    Model & \STI{} & \SIT{} & NatBench & POPE & Wino\\
    \midrule
    Qwen3-VL-8B   & 0.270 & 0.351 & $+0.081$ & $+0.021$ & $\mathbf{+0.095}$\\
    Qwen2.5-VL-7B & 0.102 & 0.277 & $+0.175$ & $+0.057$ & $\mathbf{+0.185}$\\
    InternVL3-8B  & 0.276 & 0.355 & $+0.079$ & $+0.053$ & $\mathbf{+0.210}$\\
    LLaVA-1.5-7B  & 0.000 & 0.123 & $+0.123$ & $\mathbf{+0.371}$ & $+0.048$\\
    Gemma-3-27B   & 0.232 & 0.226 & $-0.006$ & $+0.005$ & $-0.005$\\
    \bottomrule
  \end{tabular}
\end{table}

If seeing the question first informs visual encoding, why does it hurt?
Two families of explanation come to mind, and both turn out to be wrong:
perhaps the steering simply does not happen (the image tokens are encoded
the same regardless of what precedes them), or perhaps it happens and is
harmful (question-conditioned encoding discards evidence the answer
needs). \Cref{sec:analysis} rejects both. The steering happens, it is
benign, and it is thrown away.

\begin{finding}
\textbf{Finding.} Question \emph{position}, not presence, drives a large
accuracy gap: with identical content, question-first trails question-last
by 8.1 NaturalBench and 9.5 Winoground group-accuracy points on
Qwen3-VL-8B.
\end{finding}

\section{Anatomy of the Paradox}
\label{sec:analysis}

We open the model with two probes on the orderings that define the
paradox, question-first (\STI{}) and question-last (\SIT{}). The
\emph{perception probe} asks what the image tokens encode, decoding each
patch's hidden state with the logit lens and measuring how far it moves
from an image-only baseline. The \emph{read-out probe} asks what the answer
position uses, tracking layer by layer its attention over the question and
image spans and the emergence of the correct answer token. The paradox
resolves into a clean dissociation: steering lives in the first probe, the
failure in the second. Unless noted, the correlational per-layer probes run on NaturalBench
\emph{disagreement pairs}: 150 yes/no questions that question-last answers
correctly and question-first answers wrong (\cref{sec:supp-setup}).

\subsection{Perception probe: steering is real and benign}
\label{sec:steering}

Under \STI{}, image patches decode to more question-relevant concepts than
under a question-last ordering (\cref{fig:steering}). In the example the
question asks whether an adult is chasing a child; question-first turns generic
patch decodings into question-aligned ones (\emph{adult}, \emph{chase},
\emph{child}). The attention probe agrees (\cref{fig:mechanism}, middle):
under \STI{} the answer position places \emph{more} mass on the image
span than question-last does (peak 0.237 vs.\ 0.094 for \SIT{}). By the naive hypothesis
this should help. It does not: \STI{} is the worst ordering in every
table in this paper, and the steered model still answers wrong.

We can measure the steering directly (\cref{fig:cosine}). Relative to the
image shown alone, question-last (\SIT{}) barely moves the patches (mean
cosine 0.91); the small residual is the system prompt that precedes the
image, not the question, which \SIT{} places after it and which the causal
mask therefore hides from the patches. Question-first (\STI{}) moves them
markedly (mean 0.86, down to 0.62 in the most affected patches), a
spatially structured shift. A question before the image reshapes
perception; a question after it cannot.

\begin{figure}[t]
  \centering
  \includegraphics[width=\linewidth]{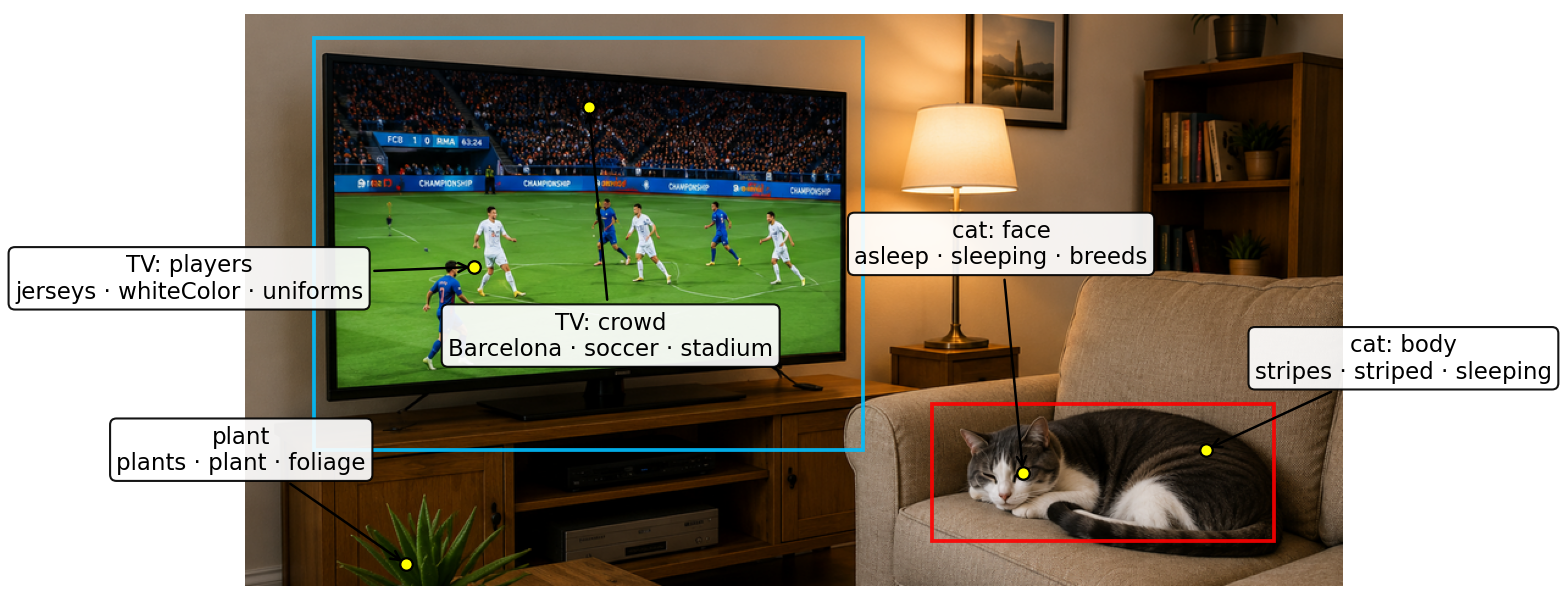}
  \caption{\textbf{The visual representation carries localized, readable object
  identity.} Per-patch logit-lens decodings (Qwen3-VL, layer 28): each region
  decodes to its own object (crowd to \emph{soccer}, cat to
  \emph{asleep}/\emph{striped}, plant to \emph{foliage}). Perception is correct
  and localized, so the question-first paradox is a downstream read-out failure
  (full analysis in \cref{fig:qq-logitlens-text}).}
  \label{fig:perception}
\end{figure}

This steered perception is also correct and well localized: decoded through
the logit lens at a late layer, every region reports its own object
(\cref{fig:perception}). The representation the read-out stage must consult is
right and localized, so the paradox is a failure to consult it, not to form it.

\subsection{Read-out probe: the question is under-attended and the
decision locks in early}
\label{sec:readout}

\begin{figure}[t]
  \centering
  \includegraphics[width=\linewidth]{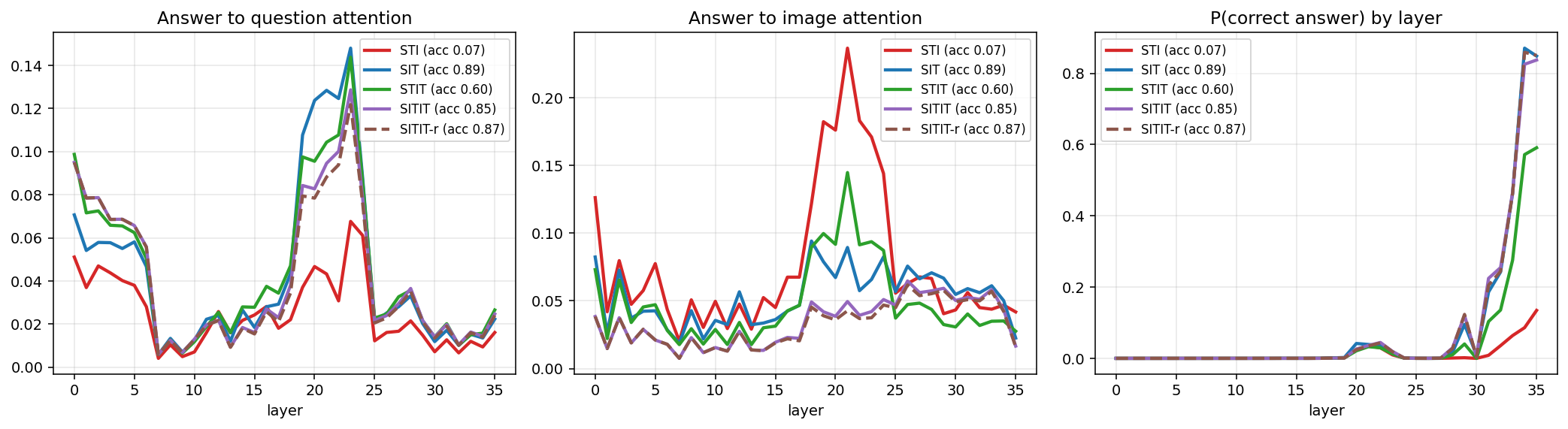}
  \caption{Per-layer probe on NaturalBench disagreement pairs
  (\STI{}-wrong, question-last-right; 150 pairs), averaged over pairs.
  \emph{Left:} answer-to-question attention; \STI{} (red) barely attends
  to the far-away question, while \SIT{}, \STIT{}, and \SITIT{} do.
  \emph{Middle:} answer-to-image attention; \STI{} over-attends to the
  image. \emph{Right:} logit-lens P(correct answer token) by layer; the
  correct answer emerges only for the orderings that place a question
  next to the answer. The re-presentation variants of \cref{sec:method}
  (\SITIT{}, dashed \SITITr{}) track echoing on these probes.}
  \label{fig:mechanism}
\end{figure}

The read-out probe resolves the paradox (\cref{fig:mechanism}, left). The
answer's attention to the question peaks at only 0.068 under \STI{} against
0.148 under \SIT{}: separated from the answer by the whole image span, the
question is barely read. The emergence curve (\cref{fig:mechanism}, right)
tells the downstream story: the probability of the correct answer token
rises with depth under question-last (final 0.85) but stays flat and low
under question-first (final 0.14).

The decision-layer view sharpens this into timing. Tracking the two-way
P(correct) over depth, \STI{} commits to a wrong, image-anchored token in
41\% of these pairs against 2\% under \SIT{}: question-first decides early,
from the image, before a question representation reaches the answer.

\begin{finding}
\textbf{Finding.} Question-first fails at read-out, not perception. The
answer under-attends the far-away question (peak 0.068 vs.\ 0.148 for
question-last) and locks onto a wrong image-anchored token in 41\% of
disagreement pairs (vs.\ 2\% for question-last); question position, not
presence, decides what the answer is read from.
\end{finding}

\subsection{Causal intervention: the read-out edge is necessary}
\label{sec:causal}
The probes above are correlational: they show the answer \emph{attends
less} to the question under \STI{}, not that this edge \emph{causes} the
gap. We test causation with an attention knockout. During a single
forward pass we add a pre-softmax $-\infty$ bias to every attention edge
from the answer position to a chosen token span, at all 36 text layers,
severing the answer's ability to read that span while leaving every other
computation intact. On 250 outcome-independent yes/no pairs we knock out,
in turn, the \emph{question} span, the \emph{image} span, and a same-size
\emph{random} text span (a specificity control), under each ordering
(\cref{tab:knockout}). The knockout leaves the question in the prompt and
cuts only the answer position's \emph{direct} edge to it: the question is
still encoded and still read by every other position, and its content can
still reach the answer indirectly through the tokens between the question
and the answer, which have already attended to it. The knockout thus
isolates the answer's direct read-out of the question, not its total
access, which is why severing an edge the answer does use lowers accuracy
only partially rather than to chance (\cref{sec:supp-knockout}).

The result is a clean double dissociation. Under question-last (\SIT{}),
cutting the answer's edge to the question lowers accuracy by $0.056$
(95\% CI $[0.020,0.096]$, paired bootstrap $p\approx0.003$), while
cutting its edge to the image does nothing ($0.000$, n.s.): the answer is
causally reading the question. Under question-first (\STI{}) the
dependence flips: the question edge is inert ($+0.004$, n.s.) and the
\emph{image} edge is what matters ($0.096$, $[0.040,0.152]$,
$p\approx0.001$). The random-span control is null in both orderings
($\le0.008$), and both interaction terms are significant (question:
$0.060$, $p\approx0.006$; image: $0.096$, $p\approx0.001$). Widening the
severed region to the whole post-question span reproduces the image
dependence more starkly (\STI{} image knockout $-0.204$). Question
position does not merely change where the model \emph{looks}; it changes
what the answer is \emph{computed from}.

\begin{table}[tb]
  \caption{Causal attention knockout (Qwen3-VL-8B, 250 yes/no pairs).
  Change in per-pair accuracy when the answer's attention to a span is
  severed at all layers (clean acc: \STI{} 0.576, \SIT{} 0.588); negative
  means the edge was used. Under question-last the answer causally reads
  the \emph{question}; under question-first, the \emph{image}.
  $^{*}$: paired-bootstrap 95\% CI excludes zero.}
  \label{tab:knockout}
  \centering
  \small
  \begin{tabular}{@{}lcc@{}}
    \toprule
    Severed edge & \STI{} (Q-first) & \SIT{} (Q-last)\\
    \midrule
    answer\,$\to$\,question & $+0.004$          & $\mathbf{-0.056}^{*}$\\
    answer\,$\to$\,image    & $\mathbf{-0.096}^{*}$ & $+0.000$\\
    answer\,$\to$\,random   & $+0.004$          & $-0.008$\\
    \bottomrule
  \end{tabular}
\end{table}

\begin{finding}
\textbf{Finding.} The read-out edge is causal, not just correlational.
Severing the answer's attention to the question costs $0.056$ accuracy under
question-last ($p\approx0.003$) but nothing under question-first, where the
image edge carries the answer instead ($0.096$, $p\approx0.001$); same-size
random knockouts are null. Question position controls what the answer is
computed from.
\end{finding}

\subsection{The mechanism predicts: the gap scales with question-answer
distance}
\label{sec:scaling}
The read-out account makes a quantitative prediction. Under \STI{} the
image span separates the question from the answer, so the more vision
tokens the image occupies, the further the question sits and the less it
is read; under \SIT{} the question is adjacent regardless of image size.
The \STI{}-\SIT{} gap should therefore widen with image tokens. We test
this on Qwen3-VL by re-rendering NaturalBench at a range of resolutions,
which sets the vision-token count, using only images large enough that
every point is a genuine downscale. The prediction holds: the gap widens
from 0.05 at 64 tokens to 0.08 at 324, then plateaus as question-last
saturates near 0.35 once the image is legible while question-first keeps
climbing (\cref{sec:supp-scaling}). The diagnosis predicts the direction
and the saturation, not merely the sign.

Together, the two probes replace the paradox with a mechanism.
Question-first prompting buys perception steering at the price of decoder
access, and in current VLMs the access matters more. The ideal prompt
should not choose between the two: it should supply the question twice.

\section{From Diagnosis to Prompt: Echoing}
\label{sec:method}

The diagnosis prescribes the prompt. Steering requires a question
\emph{before} the image (\cref{sec:steering}); read-out requires a
question \emph{adjacent to the answer} (\cref{sec:readout}). No single
question copy can be in both places, so we use two.

\subsection{Question echoing (\STIT{})}
Our base prompt restates the question on both sides of the image:
\begin{equation}
\STIT{}\;=\;\text{System},\ \text{Task},\ \text{Image},\ \text{Task}.
\end{equation}
The pre-image copy provides the top-down signal that steers visual
encoding (the promise of question-first). The post-image copy restores
question--answer adjacency (the strength of question-last). Echoing adds
only the token cost of one repeated question, a dozen tokens against
hundreds of image tokens.

The \cref{sec:analysis} probes confirm echoing realizes both halves of the
diagnosis. Its pre-image copy reproduces question-first's perceptual
steering exactly: the image sees the same tokens before it, so under the
causal mask its patch encodings coincide with \STI{}'s (\cref{fig:cosine},
cosine 1.000), and the two decode identically (\cref{fig:steering}). Its
post-image copy restores read-out: the answer now attends to the adjacent
question (\cref{fig:mechanism}; peak 0.144, against \STI{}'s 0.068), the
correct answer re-emerges with depth (final P(correct) 0.59, against 0.14),
and the model commits early to a wrong token in only 17\% of the
disagreement pairs, against \STI{}'s 41\%. Echoing keeps the steering of
question-first and the read-out of question-last.

\subsection{Echoing the image (\SITIT{}, \SITITr{})}
\label{sec:bidir}
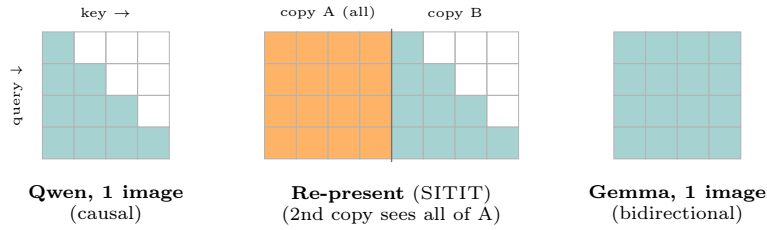
\begin{figure}[t]
  \centering
  \begin{tikzpicture}[scale=0.42, font=\scriptsize,
      cell/.style={draw=black!30, line width=0.2pt}]
    \begin{scope}[shift={(0,0)}]
      \foreach \r in {1,...,4}{\foreach \c in {1,...,4}{
        \ifnum\c<\numexpr\r+1\relax \fill[teal!35] (\c-1,-\r) rectangle (\c,-\r+1);\fi
        \draw[cell] (\c-1,-\r) rectangle (\c,-\r+1);}}
      \node[font=\tiny,anchor=south] at (2,0.12) {key $\rightarrow$};
      \node[font=\tiny,rotate=90,anchor=south] at (-0.35,-2) {query $\downarrow$};
      \node[align=center,anchor=north,font=\scriptsize] at (2,-4.55)
        {\textbf{Qwen, 1 image}\\(causal)};
    \end{scope}
    \begin{scope}[shift={(7,0)}]
      \foreach \r in {1,...,4}{\foreach \c in {1,...,8}{
        \ifnum\c<5 \fill[orange!60] (\c-1,-\r) rectangle (\c,-\r+1);
        \else\ifnum\c<\numexpr\r+5\relax \fill[teal!35] (\c-1,-\r) rectangle (\c,-\r+1);\fi\fi
        \draw[cell] (\c-1,-\r) rectangle (\c,-\r+1);}}
      \draw[black!55,line width=0.5pt] (4,0.1) -- (4,-4.1);
      \node[font=\tiny,anchor=south] at (2,0.12) {copy A (all)};
      \node[font=\tiny,anchor=south] at (6,0.12) {copy B};
      \node[align=center,anchor=north,font=\scriptsize] at (4,-4.55)
        {\textbf{Re-present} (\textsc{SITIT})\\(2nd copy sees all of A)};
    \end{scope}
    \begin{scope}[shift={(18,0)}]
      \foreach \r in {1,...,4}{\foreach \c in {1,...,4}{
        \fill[teal!35] (\c-1,-\r) rectangle (\c,-\r+1);
        \draw[cell] (\c-1,-\r) rectangle (\c,-\r+1);}}
      \node[align=center,anchor=north,font=\scriptsize] at (2,-4.55)
        {\textbf{Gemma, 1 image}\\(bidirectional)};
    \end{scope}
  \end{tikzpicture}
  \caption{\textbf{The attention reachability the mechanism turns on.} A
  filled cell means the query patch (row) can attend to the key patch
  (column). \emph{Left:} in Qwen's causal decoder a single image lets
  patch $k$ attend only to $1{\dots}k$, so its representation misses later
  patches. \emph{Middle:} re-presenting the image (\SITIT{}) makes every
  copy-B patch attend over all of copy A (orange), so it integrates the
  whole image, a bidirectional read produced by repetition alone.
  \emph{Right:} Gemma's decoder already attends bidirectionally within an
  image block, so repetition adds less. Reversing copy B (\SITITr{}) keeps
  the orange whole-image read and only flips which copy-B neighbours
  (teal) refine each patch, which is why it changes little.}
  \label{fig:attn}
\end{figure}

A stronger variant \emph{echoes the image} as well, re-presenting it
alongside the repeated question:
\begin{equation}
\SITIT{}\;=\;\text{System},\ \text{Image},\ \text{Task},\ \text{Image},\ \text{Task}.
\end{equation}
The model sees the (image, question) pair twice and answers adjacent to
the second question. Two effects combine. First, the second pass
re-encodes the image with the question already in context, yielding
question-specific features. Second, it removes a causal-attention
bottleneck. The vision encoder is bidirectional over patches, but the
Qwen3-VL decoder attends to those patch tokens causally, so within the
decoder a patch cannot draw on patches that follow it (\cref{fig:attn},
left). Repetition alone removes the limitation: under the causal mask,
every patch of the second copy attends over the \emph{entire} first copy
(\cref{fig:attn}, middle), so its representation integrates the whole
image, a bidirectional-style read obtained without changing the
architecture.

Read as a whole, \SITIT{} combines the strengths of both baselines: its
opening image-question pair reproduces \SIT{}, the trained format,
encoded in-distribution, and its second image is the question-steered
pass of \STI{} (\cref{fig:cosine}), now attending over the whole first
copy and sitting adjacent to the answer, where the decoder can read it
instead of stranding it.

A further variant reverses the second copy's patch order and positional
indices (\SITITr{}); \cref{sec:reversal} shows its effect is small and
follows the decoder's attention mask.

\section{Results}
\label{sec:exp}

\subsection{The position ladder}
\begin{figure}[tb]
  \centering
  \includegraphics[width=0.9\linewidth]{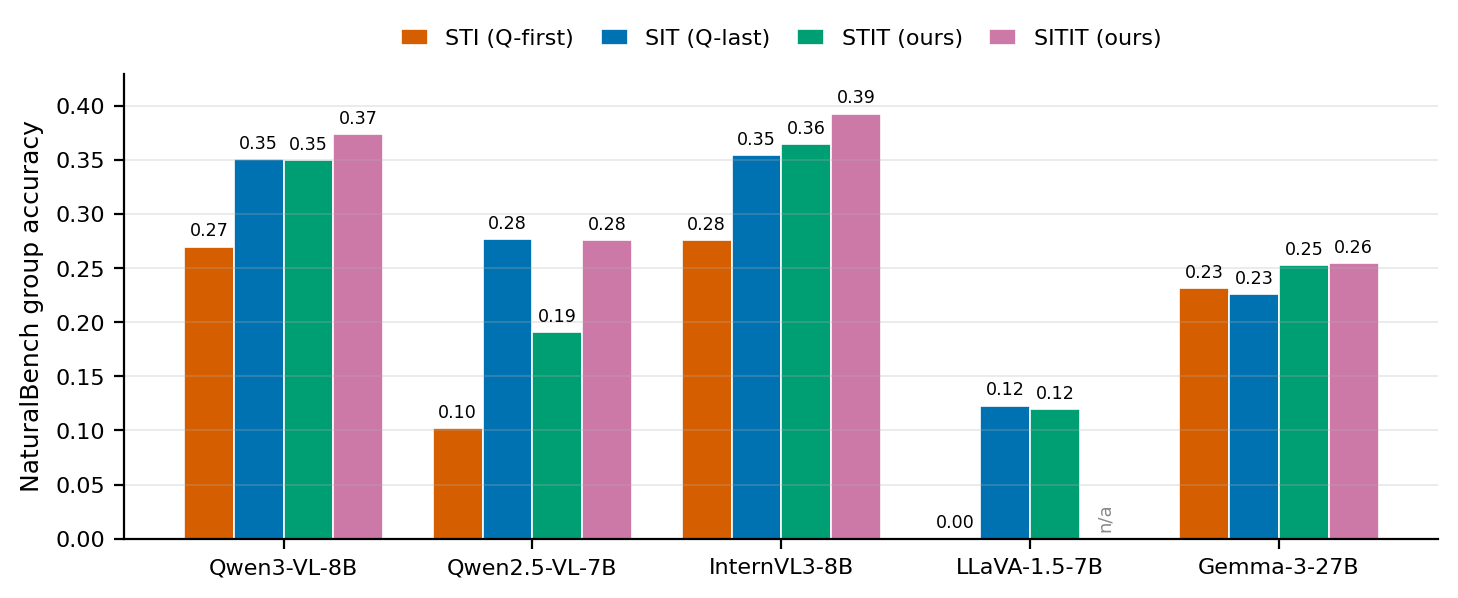}
  \caption{\textbf{The position ladder, across models.} NaturalBench group
  accuracy for the five VLMs. On the four models with a paradox, accuracy
  climbs from question-first (\STI{}) to question-last (\SIT{}) to echoing
  (\STIT{}) to image re-presentation (\SITIT{}); the exceptions are honest and
  visible (Qwen2.5-VL under-recovers at \STIT{}; Gemma-3 has no \STI{}-\SIT{}
  gap but still gains from echoing; LLaVA-1.5 is single-image, so \SITIT{} is
  n/a). Exact per-metric numbers for all models are in
  \cref{tab:naturalbench-full}.}
  \label{fig:ladder}
\end{figure}

\Cref{fig:ladder} plots the NaturalBench position ladder across all five
models, and \cref{tab:main} adds POPE and Winoground. The results form a
ladder. On Qwen3-VL, moving the question after the image recovers 8.1
NaturalBench group-accuracy points with the same tokens; echoing matches
that question-last baseline, and re-presenting the image is best, a
significant $+0.024$ over echoing and 10.4 points over question-first, with
every sub-metric in the same order (\cref{tab:naturalbench-full}). The
pattern holds across models, with two honest exceptions the figure makes
visible. On Gemma-3, where question order does not matter, echoing still
helps significantly on NaturalBench and Winoground (near-saturated POPE stays
flat), so the fix is not merely undoing a Qwen-family quirk. On Qwen2.5-VL, question echoing alone under-recovers, but image
re-presentation restores it. LLaVA-1.5 is the extreme case: question-first
collapses to a constant ``Yes'' on every benchmark and question-last
repairs it, its POPE gap ($+0.371$) the largest of any model; being
single-image, it has no image-re-presentation ordering.

The ladder is steepest on compositional Winoground: Qwen3-VL climbs from
0.223 to 0.410 group accuracy along the ladder, a 19-point gain over
question-first ($p<10^{-13}$), and Gemma-3 from 0.213 to 0.318. On
near-saturated POPE every ordering except question-first sits within a
point of the best, so we do not over-read it.

The two echoed variants carry different costs and different wins. Question
echoing (\STIT{}), which adds no image tokens, recovers the gap and beats
question-last on NaturalBench and Winoground, but only matches it on
near-saturated POPE and slightly trails it on VQAv2 (\cref{sec:openended}); its
clean win is the compositional benchmarks. Image re-presentation (\SITIT{}) is
uniformly best but adds a second image, roughly doubling image prefill
(\cref{sec:cost}). One model resists the free fix: on Qwen2.5-VL, \STIT{} alone
under-recovers and only the image copy reaches question-last
(\cref{tab:naturalbench-full}). The free variant is a strong default, not a
universal one.

\begin{table}[tb]
  \caption{\textbf{Main results (primary models).} NaturalBench group
  accuracy (1{,}900 groups), POPE accuracy/F1 (9{,}000 questions), and
  Winoground text/group accuracy (400 groups) for the two primary models.
  Question echoing (\STIT{}) recovers the question-first gap and image
  re-presentation (\SITIT{}, \SITITr{}) matches or exceeds question-last,
  winning most on the hardest benchmark (Winoground). On Qwen3-VL the
  \STI{}$\to$\SIT{} gap and the \STIT{}$\to$\SITIT{} gain are significant
  (McNemar $p<10^{-12}$, $p=0.016$); on Gemma \STI{}$\approx$\SIT{}
  ($p=0.62$) yet echoing still helps. Best per model in \textbf{bold}.
  \Cref{fig:ladder} shows the NaturalBench ladder for all five models.}
  \label{tab:main}
  \centering
  \small
  \begin{tabular}{@{}llccccc@{}}
    \toprule
    & & NatBench & \multicolumn{2}{c}{POPE} & \multicolumn{2}{c}{Winoground}\\
    \cmidrule(lr){3-3}\cmidrule(lr){4-5}\cmidrule(lr){6-7}
    Model & Ordering & Group & Acc & F1 & Text & Group\\
    \midrule
    \multirow{5}{*}{Qwen3-VL-8B}
      & \STI{} (Q-first) & 0.270 & 0.870 & 0.861 & 0.699 & 0.223\\
      & \SIT{} (Q-last)  & 0.351 & \textbf{0.891} & \textbf{0.884} & 0.743 & 0.318\\
      & \STIT{} (ours)   & 0.350 & 0.884 & 0.876 & 0.774 & 0.375\\
      & \SITIT{} (ours)  & \textbf{0.374} & 0.889 & 0.881 & 0.779 & 0.403\\
      & \SITITr{} (ours) & \textbf{0.374} & 0.889 & 0.883 & \textbf{0.781} & \textbf{0.410}\\
    \midrule
    \multirow{5}{*}{Gemma-3-27B}
      & \STI{} (Q-first) & 0.232 & 0.840 & 0.830 & 0.680 & 0.213\\
      & \SIT{} (Q-last)  & 0.226 & \textbf{0.845} & \textbf{0.845} & 0.683 & 0.208\\
      & \STIT{} (ours)   & 0.253 & 0.838 & 0.832 & 0.708 & 0.253\\
      & \SITIT{} (ours)  & \textbf{0.255} & 0.834 & 0.836 & \textbf{0.733} & \textbf{0.318}\\
      & \SITITr{} (ours) & 0.252 & 0.835 & 0.837 & 0.721 & 0.298\\
    \bottomrule
  \end{tabular}
\end{table}

\begin{finding}
\textbf{Finding.} The gains are largest where the task is hardest. On
Winoground group accuracy our prompts beat the best baseline by 9.2 points
on Qwen3-VL (0.410 vs.\ 0.318, $p<10^{-5}$) and 11.0 on Gemma-3 (0.318 vs.\
0.208, $p<10^{-6}$); on near-saturated POPE all orderings except
question-first are within a point.
\end{finding}

\paragraph{Open-ended generation.}
\label{sec:openended}
Every benchmark so far is yes/no or forced-choice, so one might worry the
ladder is an artifact of a constrained output space. It is not: on
2{,}000 VQAv2 validation questions scored with the official soft-accuracy
metric over free-form generation, the ladder survives (question-first
0.811, question-last 0.834, image re-presentation best at 0.838; the
\STI{}-\SIT{} gap excludes zero at 95\%; \cref{tab:vqa}). On this open-ended
metric question echoing alone recovers part of the gap (0.821) but does not
surpass question-last; the win over question-last comes from re-presenting the
image. The paradox itself is a property of where the question sits, not of the
answer format.

\begin{table}[t]
  \caption{Open-ended VQA (Qwen3-VL-8B, 2{,}000 VQAv2 validation
  questions, official soft-accuracy over free-form generation). The
  position ladder survives unconstrained output: question-first trails
  question-last (paired bootstrap $\Delta=+0.023$, 95\% CI
  $[+0.011,+0.035]$) and image re-presentation is best. Best in
  \textbf{bold}.}
  \label{tab:vqa}
  \centering
  \small
  \begin{tabular}{@{}lcccc@{}}
    \toprule
    & \STI{} & \SIT{} & \STIT{} & \SITIT{}\\
    \midrule
    VQA soft-acc & 0.811 & 0.834 & 0.821 & \textbf{0.838}\\
    \bottomrule
  \end{tabular}
\end{table}

\subsection{The cost of each ordering}
\label{sec:cost}
The two echoed variants carry very different inference cost, a deployment fact
worth stating. A VLM prefills the whole prompt before decoding, so cost tracks
the token count: the KV-cache grows linearly and self-attention prefill
quadratically. Image tokens dominate (\cref{tab:cost}: 2{,}305 per image against
about 65 for system-plus-question on Qwen3-VL), so the variants sit at opposite
ends. Question echoing (\STIT{}) adds only a second question, about 27 tokens
(+1\%), leaving cost essentially unchanged; image re-presentation (\SITIT{}) adds
a second image and nearly doubles the prefill (1.98$\times$ cache, 3.9$\times$
attention FLOPs). Accuracy and cost rank the same way: \STIT{} is the near-free
default, and \SITIT{} buys its extra accuracy with double the prefill. Image
tokens dominate on every VLM we tested, so this picture is general.

\begin{table}[tb]
  \caption{\textbf{Prompt cost per ordering} (Qwen3-VL-8B, mean over 30
  NaturalBench prompts). A VLM prefills the full prompt, so cost tracks token
  count: KV-cache scales linearly, self-attention prefill quadratically.
  Question echoing (\STIT{}) is within 1\% of the single-pass orderings; image
  re-presentation (\SITIT{}) roughly doubles the prefill.}
  \label{tab:cost}
  \centering
  \small
  \begin{tabular}{@{}lccccc@{}}
    \toprule
    Ordering & Image tok & Text tok & Total & KV-cache & Attn.\ prefill\\
    \midrule
    \STI{} / \SIT{} & 2{,}305 & 65 & 2{,}370 & $1.0\times$ & $1.0\times$\\
    \STIT{} (ours)  & 2{,}305 & 93 & 2{,}398 & $1.01\times$ & $1.02\times$\\
    \SITIT{} (ours) & 4{,}610 & 96 & 4{,}706 & $1.98\times$ & $3.94\times$\\
    \bottomrule
  \end{tabular}
\end{table}

\section{Ablations: Isolating the Active Ingredient}
\label{sec:ablation}

The mechanism claims the gap is about \emph{where the question sits}, not
about token count or compute. We test the claim by trying to close the gap
with everything except a post-image question (\cref{tab:ablation};
Qwen3-VL-8B, NaturalBench group accuracy).

\begin{table}[tb]
  \caption{Ablations on NaturalBench (Qwen3-VL-8B, group accuracy).
  Adding image tokens or pre-answer compute does not close the
  question-first gap; a post-image question does.}
  \label{tab:ablation}
  \centering
  \small
  \begin{tabular}{@{}llc@{}}
    \toprule
    Group & Variant & Group acc\\
    \midrule
    \multirow{3}{*}{Not more tokens}
      & \STI{} (baseline)          & 0.270\\
      & \STI{} $+$ img copies $\times2/\times3$ & 0.281 / 0.280\\
      & \STI{} $+$ mean-resize (fewer) & 0.258\\
    \midrule
    \multirow{2}{*}{Not more compute}
      & \STI{} $+$ space (5/40/120 pad) & 0.266 / 0.267 / 0.266\\
      & \STIT{} + chain-of-thought & 0.332\\
    \midrule
    \multirow{2}{*}{Question content matters}
      & \STIT{} (full post-image question) & 0.350\\
      & \STIT{} + short cue (same position) & 0.292\\
    \midrule
    \multirow{6}{*}{Ordering ladder}
      & \SIT{} (question-last)                       & 0.351\\
      & \STIT{} (echo question)                      & 0.350\\
      & \textsc{STITI} (echo image, \STI{} units)   & 0.354\\
      & \textsc{STITIT} (\textsc{STITI} $+$ final Q) & 0.352\\
      & \SITIT{} (echo image, \SIT{} units)         & \textbf{0.374}\\
      & \SITITr{} (\SITIT{}, 2nd image rev.)        & \textbf{0.374}\\
    \bottomrule
  \end{tabular}
\end{table}

\paragraph{It is not more image tokens.} Duplicating the image under
\STI{} lifts group accuracy only marginally (to about 0.28), and
shrinking the image to fewer tokens (mean-resize) lowers it to 0.258.

\paragraph{It is not more pre-answer compute.} Inserting 5 to 120 padding
tokens after the question does not help (about 0.266), and
chain-of-thought on top of echoing does not add either (0.332 vs.\
0.350).

\paragraph{The question content is what must be adjacent.} A full
post-image question restores accuracy (\STIT{} 0.350); a short cue in the
same position recovers far less (0.292). The post-image copy must restate
the question, not merely mark where it ended.

\paragraph{Repeating the image helps only in the training order.} The
natural way to extend question echoing with a repeated image is
\textsc{STITI}: echo the question around the image, then repeat the
image. It underperforms \SITIT{} (0.354 vs.\ 0.374) even though the two
contain \emph{identical} content and differ only in how each image and
question are paired: \SITIT{} uses the image-first units of \SIT{}, the
format the model is trained with, whereas \textsc{STITI} uses the
question-first units of \STI{}. The gap is not adjacency: appending a
final question to \textsc{STITI} (\textsc{STITIT}) does not help (0.352).
Re-presenting the image pays off only when the repeated units follow the
in-distribution image-first order.

\begin{finding}
\textbf{Finding.} The lever is a restated question adjacent to the
answer, not token count or compute: more image tokens (0.281), fewer
(0.258), and padding (0.266) leave the gap open, while a full post-image
question closes it (0.350; a short cue does not, 0.292). Re-presenting
the image helps only in the training format: \SITIT{} beats
content-identical \textsc{STITI} by two points (0.374 vs.\ 0.354).
\end{finding}

\section{Discussion}
\label{sec:discussion}

\paragraph{Two mechanisms, two positions.} Our results separate two
effects that prompt design conflates. Putting the question first does
create top-down visual steering, but a decoder-only VLM reads its answer
mostly from tokens adjacent to the answer position, so a stranded
question is barely consulted and the steered features are decoded into an
early, image-anchored guess. Question-last wins not because it steers
vision better (it steers less) but because the question sits where the
answer can use it. Echoing gets both: steer with the pre-image copy, read
out with the post-image copy.

\paragraph{Implications beyond prompting.} Three follow. \emph{For
benchmarking:} reported VQA numbers depend on an undocumented ordering choice worth
up to 17.5 points; protocols should fix and report it. \emph{For agentic
systems:} pipelines that state the task before attaching visual context run
the worst ordering, and a one-line echo repairs them. \emph{For training:}
the in-distribution effect (\cref{sec:ablation}) suggests order-robustness is
learned; mixing section orders during instruction-tuning may remove the
paradox at the source.

\paragraph{Limitations.} We study open VLMs up to 27B parameters on
benchmarks dominated by short-answer questions; closed models and
long-form generation beyond VQAv2 remain future work. The logit lens is
approximate, and activation patching would localize the circuit beyond
our knockout. Re-presentation adds the token cost of a second image.
Gemma-3-27B, our only model without the paradox, is also our largest and our
only one with bidirectional image attention, so we cannot say whether the
paradox fades with scale, with bidirectional attention, or both; its
bidirectional attention and pretraining may already blunt the read-out failure
that echoing targets. Echoing helps it regardless; separating these causes needs
models that vary scale and attention independently.

\paragraph{Conclusion.} The question-first paradox is a positional
read-out failure: the ordering that best steers perception strands the question where the
answer cannot read it. The fix is a prompt edit: put the question where it can
both steer and be read, and give the decoder a second look at the image.

\bibliographystyle{splncs04}
\bibliography{references}

@String(ICLR  = {Int. Conf. Learn. Represent.})

@String(ICLR  = {ICLR})

@article{AprilGemini,
    title = {{April 2025 Edition Vertex AI Gemini}},
    url = {https://services.google.com/fh/files/misc/2_vertex_ai_gemini_multimodal_prompting.pdf}
}

@article{Shaier2024AskingQuestionsb,
    title = {{Asking Again and Again: Exploring LLM Robustness to Repeated Questions}},
    year = {2024},
    author = {Shaier, Sagi and Sanz-Guerrero, Mario and Von Der Wense, Katharina},
    month = {12},
    url = {https://arxiv.org/pdf/2412.07923},
    arxivId = {2412.07923}
}

@article{Li2023EvaluatingModels,
    title = {{Evaluating Object Hallucination in Large Vision-Language Models}},
    year = {2023},
    journal = {EMNLP 2023 - 2023 Conference on Empirical Methods in Natural Language Processing, Proceedings},
    author = {Li, Yifan and Du, Yifan and Zhou, Kun and Wang, Jinpeng and Zhao, Wayne Xin and Wen, Ji Rong},
    month = {5},
    pages = {292--305},
    publisher = {Association for Computational Linguistics (ACL)},
    url = {https://arxiv.org/pdf/2305.10355},
    isbn = {9798891760608},
    doi = {10.18653/v1/2023.emnlp-main.20},
    arxivId = {2305.10355}
}

@article{Team2025GemmaReport,
    title = {{Gemma 3 Technical Report}},
    year = {2025},
    author = {Team, Gemma and Kamath, Aishwarya and Ferret, Johan and Pathak, Shreya and Vieillard, Nino and Merhej, Ramona and Perrin, Sarah and Matejovicova, Tatiana and Ram{\'{e}}, Alexandre and Rivi{\`{e}}re, Morgane and Rouillard, Louis and Mesnard, Thomas and Cideron, Geoffrey and Grill, Jean-bastien and Ramos, Sabela and Yvinec, Edouard and Casbon, Michelle and Pot, Etienne and Penchev, Ivo and Liu, Gaël and Visin, Francesco and Kenealy, Kathleen and Beyer, Lucas and Zhai, Xiaohai and Tsitsulin, Anton and Busa-Fekete, Robert and Feng, Alex and Sachdeva, Noveen and Coleman, Benjamin and Gao, Yi and Mustafa, Basil and Barr, Iain and Parisotto, Emilio and Tian, David and Eyal, Matan and Cherry, Colin and Peter, Jan-Thorsten and Sinopalnikov, Danila and Bhupatiraju, Surya and Agarwal, Rishabh and Kazemi, Mehran and Malkin, Dan and Kumar, Ravin and Vilar, David and Brusilovsky, Idan and Luo, Jiaming and Steiner, Andreas and Friesen, Abe and Sharma, Abhanshu and Sharma, Abheesht and Gilady, Adi Mayrav and Goedeckemeyer, Adrian and Saade, Alaa and Feng, Alex and Kolesnikov, Alexander and Bendebury, Alexei and Abdagic, Alvin and Vadi, Amit and Gy{\"{o}}rgy, András and Pinto, André Susano and Das, Anil and Bapna, Ankur and Miech, Antoine and Yang, Antoine and Paterson, Antonia and Shenoy, Ashish and Chakrabarti, Ayan and Piot, Bilal and Wu, Bo and Shahriari, Bobak and Petrini, Bryce and Chen, Charlie and Lan, Charline Le and Choquette-Choo, Christopher A. and Carey, CJ and Brick, Cormac and Deutsch, Daniel and Eisenbud, Danielle and Cattle, Dee and Cheng, Derek and Paparas, Dimitris and Sreepathihalli, Divyashree Shivakumar and Reid, Doug and Tran, Dustin and Zelle, Dustin and Noland, Eric and Huizenga, Erwin and Kharitonov, Eugene and Liu, Frederick and Amirkhanyan, Gagik and Cameron, Glenn and Hashemi, Hadi and Klimczak-Pluci{\'{n}}ska, Hanna and Singh, Harman and Mehta, Harsh and Lehri, Harshal Tushar and Hazimeh, Hussein and Ballantyne, Ian and Szpektor, Idan and Nardini, Ivan and Pouget-Abadie, Jean and Chan, Jetha and Stanton, Joe and Wieting, John and Lai, Jonathan and Orbay, Jordi and Fernandez, Joseph and Newlan, Josh and Ji, Ju-yeong and Singh, Jyotinder and Black, Kat and Yu, Kathy and Hui, Kevin and Vodrahalli, Kiran and Greff, Klaus and Qiu, Linhai and Valentine, Marcella and Coelho, Marina and Ritter, Marvin and Hoffman, Matt and Watson, Matthew and Chaturvedi, Mayank and Moynihan, Michael and Ma, Min and Babar, Nabila and Noy, Natasha and Byrd, Nathan and Roy, Nick and Momchev, Nikola and Chauhan, Nilay and Sachdeva, Noveen and Bunyan, Oskar and Botarda, Pankil and Caron, Paul and Rubenstein, Paul Kishan and Culliton, Phil and Schmid, Philipp and Sessa, Pier Giuseppe and Xu, Pingmei and Stanczyk, Piotr and Tafti, Pouya and Shivanna, Rakesh and Wu, Renjie and Pan, Renke and Rokni, Reza and Willoughby, Rob and Vallu, Rohith and Mullins, Ryan and Jerome, Sammy and Smoot, Sara and Girgin, Sertan and Iqbal, Shariq and Reddy, Shashir and Sheth, Shruti and Põder, Siim and Bhatnagar, Sijal and Panyam, Sindhu Raghuram and Eiger, Sivan and Zhang, Susan and Liu, Tianqi and Yacovone, Trevor and Liechty, Tyler and Kalra, Uday and Evci, Utku and Misra, Vedant and Roseberry, Vincent and Feinberg, Vlad and Kolesnikov, Vlad and Han, Woohyun and Kwon, Woosuk and Chen, Xi and Chow, Yinlam and Zhu, Yuvein and Wei, Zichuan and Egyed, Zoltan and Cotruta, Victor and Giang, Minh and Kirk, Phoebe and Rao, Anand and Black, Kat and Babar, Nabila and Lo, Jessica and Moreira, Erica and Martins, Luiz Gustavo and Sanseviero, Omar and Gonzalez, Lucas and Gleicher, Zach and Warkentin, Tris and Mirrokni, Vahab and Senter, Evan and Collins, Eli and Barral, Joelle and Ghahramani, Zoubin and Hadsell, Raia and Matias, Yossi and Sculley, D. and Petrov, Slav and Fiedel, Noah and Shazeer, Noam and Vinyals, Oriol and Dean, Jeff and Hassabis, Demis and Kavukcuoglu, Koray and Farabet, Clement and Buchatskaya, Elena and Alayrac, Jean-Baptiste and Anil, Rohan and {Dmitry} and {Lepikhin} and Borgeaud, Sebastian and Bachem, Olivier and Joulin, Armand and Andreev, Alek and Hardin, Cassidy and Dadashi, Robert and Hussenot, Léonard},
    month = {3},
    url = {https://arxiv.org/pdf/2503.19786},
    arxivId = {2503.19786}
}

@book{Buswell1935HowArt,
    title = {{How People Look at Pictures: A Study of the Psychology of Perception in Art}},
    year = {1935},
    author = {Buswell, Guy Thomas},
    publisher = {University of Chicago Press}
}

@misc{ImageLearn,
    title = {{Image prompt engineering techniques - Microsoft Foundry | Microsoft Learn}},
    url = {https://learn.microsoft.com/en-us/azure/foundry/openai/concepts/gpt-4-v-prompt-engineering}
}

@misc{2020InterpretingLessWrong,
    title = {{interpreting GPT: the logit lens — LessWrong}},
    year = {2020},
    url = {https://www.lesswrong.com/posts/AcKRB8wDpdaN6v6ru/interpreting-gpt-the-logit-lens}
}

@article{Liu2023LostContexts,
    title = {{Lost in the Middle: How Language Models Use Long Contexts}},
    year = {2023},
    journal = {Transactions of the Association for Computational Linguistics},
    author = {Liu, Nelson F. and Lin, Kevin and Hewitt, John and Paranjape, Ashwin and Bevilacqua, Michele and Petroni, Fabio and Liang, Percy},
    month = {7},
    pages = {157--173},
    volume = {12},
    publisher = {MIT Press Journals},
    url = {https://arxiv.org/pdf/2307.03172},
    isbn = {2307.03172v3},
    doi = {10.1162/tacl{\_}a{\_}00638},
    issn = {2307387X},
    arxivId = {2307.03172}
}

@article{Li2024NaturalBench:Samples,
    title = {{NaturalBench: Evaluating Vision-Language Models on Natural Adversarial Samples}},
    year = {2024},
    journal = {Advances in Neural Information Processing Systems},
    author = {Li, Baiqi and Lin, Zhiqiu and Peng, Wenxuan and de Dieu Nyandwi, Jean and Jiang, Daniel and Ma, Zixian and Khanuja, Simran and Krishna, Ranjay and Neubig, Graham and Ramanan, Deva},
    month = {10},
    volume = {37},
    publisher = {Neural information processing systems foundation},
    url = {https://arxiv.org/pdf/2410.14669},
    isbn = {2410.14669v4},
    doi = {10.52202/079017-0542},
    issn = {10495258},
    arxivId = {2410.14669}
}

@article{Tan2024OrderModels,
    title = {{Order Matters: Exploring Order Sensitivity in Multimodal Large Language Models}},
    year = {2024},
    author = {Tan, Zhijie and Chu, Xu and Li, Weiping and Mo, Tong},
    month = {10},
    url = {https://arxiv.org/pdf/2410.16983},
    arxivId = {2410.16983}
}

@article{Leviathan2025PromptLLMs,
    title = {{Prompt Repetition Improves Non-Reasoning LLMs}},
    year = {2025},
    author = {Leviathan, Yaniv and Kalman, Matan and Matias, Yossi and Research, Google},
    month = {12},
    url = {https://arxiv.org/pdf/2512.14982},
    arxivId = {2512.14982}
}

@article{Ismithdeen2025Promptception:Prompts,
    title = {{Promptception: How Sensitive Are Large Multimodal Models to Prompts?}},
    year = {2025},
    author = {Ismithdeen, Mohamed Insaf and Khattak, Muhammad Uzair and Khan, Salman},
    month = {9},
    url = {https://arxiv.org/pdf/2509.03986},
    arxivId = {2509.03986}
}

@article{Bai2025Qwen2.5-VLReportb,
    title = {{Qwen2.5-VL Technical Report}},
    year = {2025},
    author = {Bai, Shuai and Chen, Keqin and Liu, Xuejing and Wang, Jialin and Ge, Wenbin and Song, Sibo and Dang, Kai and Wang, Peng and Wang, Shijie and Tang, Jun and Zhong, Humen and Zhu, Yuanzhi and Yang, Mingkun and Li, Zhaohai and Wan, Jianqiang and Wang, Pengfei and Ding, Wei and Fu, Zheren and Xu, Yiheng and Ye, Jiabo and Zhang, Xi and Xie, Tianbao and Cheng, Zesen and Zhang, Hang and Yang, Zhibo and Xu, Haiyang and Lin, Junyang},
    month = {2},
    pages = {1--23},
    url = {https://arxiv.org/pdf/2502.13923},
    arxivId = {2502.13923}
}

@article{Xu2023Re-ReadingModels,
    title = {{Re-Reading Improves Reasoning in Large Language Models}},
    year = {2023},
    journal = {EMNLP 2024 - 2024 Conference on Empirical Methods in Natural Language Processing, Proceedings of the Conference},
    author = {Xu, Xiaohan and Tao, Chongyang and Shen, Tao and Xu, Can and Xu, Hongbo and Long, Guodong and Lou, Jian Guang and Ma, Shuai},
    month = {9},
    pages = {15549--15575},
    publisher = {Association for Computational Linguistics (ACL)},
    url = {https://arxiv.org/pdf/2309.06275},
    isbn = {9798891761643},
    doi = {10.18653/v1/2024.emnlp-main.871},
    arxivId = {2309.06275}
}

@misc{HanREADREADING,
    title = {{Read Before You Think: Mitigating LLM Comprehension Failures with Step-by-Step Reading}},
    author = {Han, Feijiang and Cui, Hengtao and Guo, Licheng and Wang, Zelong and Lyu, Zhiyuan},
    year = {2025},
    month = {4},
    url = {https://arxiv.org/abs/2504.09402},
    arxivId = {2504.09402}
}

@book{Boyd1974RepeatedLearning,
    title = {{Repeated questions in prose learning}},
    year = {1974},
    author = {Boyd, William McKinney},
    edition = {1},
    pages = {31--38},
    volume = {64},
    publisher = {American Psychological Association}
}

@article{Hamaker1986TheLearning,
    title = {{The Effects of Adjunct Questions on Prose Learning}},
    year = {1986},
    journal = {Review of Educational Research Summer},
    author = {Hamaker, Christiaan},
    number = {2},
    pages = {212--242},
    volume = {56},
    url = {http://rer.aera.net}
}

@article{Neo2024TowardsModels,
    title = {{Towards Interpreting Visual Information Processing in Vision-Language Models}},
    year = {2024},
    journal = {13th International Conference on Learning Representations, ICLR 2025},
    author = {Neo, Clement and Ong, Luke and Torr, Philip and Geva, Mor and Krueger, David and Barez, Fazl},
    month = {10},
    pages = {25461--25478},
    publisher = {International Conference on Learning Representations, ICLR},
    url = {https://arxiv.org/pdf/2410.07149},
    isbn = {9798331320850},
    arxivId = {2410.07149}
}

@article{Liu2023VisualTuningb,
    title = {{Visual Instruction Tuning}},
    year = {2023},
    journal = {Advances in Neural Information Processing Systems},
    author = {Liu, Haotian and Li, Chunyuan and Wu, Qingyang and Lee, Yong Jae},
    month = {4},
    volume = {36},
    publisher = {Neural information processing systems foundation},
    url = {https://arxiv.org/pdf/2304.08485},
    isbn = {9781713899921},
    issn = {10495258},
    arxivId = {2304.08485}
}

@article{Thrush2022Winoground:Compositionality,
    title = {{Winoground: Probing Vision and Language Models for Visio-Linguistic Compositionality}},
    year = {2022},
    journal = {Proceedings of the IEEE Computer Society Conference on Computer Vision and Pattern Recognition},
    author = {Thrush, Tristan and Jiang, Ryan and Bartolo, Max and Singh, Amanpreet and Williams, Adina and Kiela, Douwe and Ross, Candace},
    month = {4},
    pages = {5228--5238},
    volume = {2022-June},
    publisher = {IEEE Computer Society},
    url = {https://arxiv.org/pdf/2204.03162},
    isbn = {9781665469463},
    doi = {10.1109/CVPR52688.2022.00517},
    issn = {10636919},
    arxivId = {2204.03162}
}

@misc{Gupta2026TestTimeModality,
    title = {{Test-Time Training for Modality Order Consistency in Vision-Language Models}},
    author = {Gupta, Aditi and Gandelsman, Yossi},
    year = {2026},
    month = {7},
    url = {https://arxiv.org/abs/2607.20351},
    arxivId = {2607.20351}
}

@misc{QwenVL3Card,
    title = {{Qwen3-VL-8B-Instruct model card}},
    author = {{Qwen Team}},
    year = {2025},
    howpublished = {Hugging Face},
    url = {https://huggingface.co/Qwen/Qwen3-VL-8B-Instruct}
}

@misc{QwenVL25Card,
    title = {{Qwen2.5-VL-7B-Instruct model card}},
    author = {{Qwen Team}},
    year = {2025},
    howpublished = {Hugging Face},
    url = {https://huggingface.co/Qwen/Qwen2.5-VL-7B-Instruct}
}

@misc{Gemma3Card,
    title = {{gemma-3-27b-it model card}},
    author = {{Google DeepMind}},
    year = {2025},
    howpublished = {Hugging Face},
    url = {https://huggingface.co/google/gemma-3-27b-it}
}

@misc{InternVL3Card,
    title = {{InternVL3-8B model card}},
    author = {{OpenGVLab}},
    year = {2025},
    howpublished = {Hugging Face},
    url = {https://huggingface.co/OpenGVLab/InternVL3-8B}
}

@misc{LLaVA15Card,
    title = {{llava-1.5-7b-hf model card}},
    author = {{LLaVA Team}},
    year = {2023},
    howpublished = {Hugging Face},
    url = {https://huggingface.co/llava-hf/llava-1.5-7b-hf}
}

\clearpage
\appendix
\setlength{\textfloatsep}{8pt plus 3pt minus 3pt}
\setlength{\floatsep}{7pt plus 3pt minus 2pt}
\setlength{\intextsep}{7pt plus 3pt minus 2pt}
\makeatletter
\setlength{\@fptop}{0pt}
\setlength{\@fpsep}{8pt plus 1fil}
\makeatother
\section{Supplementary: experimental setup and reproducibility}
\label{sec:supp-setup}

\paragraph{Models.} We use the public instruction-tuned checkpoints
Qwen3-VL-8B, Qwen2.5-VL-7B, InternVL3-8B, LLaVA-1.5-7B, and Gemma-3-27B,
driven by a single order-aware input builder so that orderings differ only
in the arrangement of identical tokens. LLaVA-1.5 accepts a single image, so
the image-re-presentation orderings (\SITIT{}, \SITITr{}) do not apply to it.

\paragraph{Prompt.} The fixed system prompt is, verbatim:
\begin{quote}\ttfamily\small
A chat between a curious user and an artificial intelligence assistant. The
assistant gives helpful, detailed, and polite answers to the human's
questions.
\end{quote}
Yes/no questions append ``\texttt{Answer the question using only Yes or No.}'';
two-alternative items append ``\texttt{Answer with only the option letter (A or
B).}'' Decoding is greedy with a 16-token budget, and the prediction is the
first Yes/No (or option-letter) token.

\paragraph{Data.} NaturalBench ($1{,}900$ groups of four (image, question)
pairs), POPE ($9{,}000$ questions), Winoground ($400$ groups), and an
open-ended split of VQAv2 ($2{,}000$ validation questions). Group accuracy
counts a NaturalBench or Winoground group correct only if all four of its
pairs are.

\paragraph{Open-ended VQA scoring.} On VQAv2 the model generates a free-form
answer (greedy, no options given). We report the official VQA soft-accuracy:
after the standard answer normalization (lowercasing, punctuation, and
digit/article handling on both the reply and the ten human answers), a
prediction that agrees with $k$ of the ten annotators scores $\min(k/3, 1)$,
averaged over the leave-one-out annotator subsets and over questions. Because
free-form generation returns a phrase rather than a bare token, we treat a
ground-truth answer as produced if it appears in the reply as a whole word
(so ``no'' does not match ``snow''); this whole-word containment is a small
relaxation of the exact match used with short-answer models.

\paragraph{Significance.} For adjacent orderings we test the paired difference
in per-group correctness with a two-sided exact McNemar test and a paired
bootstrap $95\%$ CI ($5{,}000$ resamples).

\paragraph{Mechanistic probes.} The perception and read-out probes read a
single eager-attention forward pass per (image, question) pair; unless noted
they run on $150$ disagreement pairs (question-last correct, question-first
wrong). Disagreement pairs are chosen deliberately: they are the cases where
ordering alone changes the answer, so they act as a microscope on the mechanism,
and the per-layer separations are illustrative of it rather than
population-average effect sizes. The causal knockout that anchors our conclusion
(\cref{sec:causal}) instead uses $250$ outcome-independent yes/no pairs, so the
mechanism does not rest on this selection. Code and a project website are
included in the supplementary material.

\section{Supplementary: full per-metric results and controls}
\label{sec:supp-results}

\paragraph{Paradox significance.} \Cref{tab:significance} reports the paired
significance of every gap in \cref{tab:paradox}, plus the echoing ladder steps
on the two primary models. The question-first paradox is significant on every
model and benchmark except Gemma-3, the one model where the ordering does not
matter.

\begin{table}[t]
  \caption{Paired significance for the gaps of \cref{tab:paradox} and the
  echoing ladder. $\Delta$ is the gap on the benchmark's correctness unit
  (NaturalBench and Winoground per-group, all four (image, question) pairs
  correct; POPE per-question); the $95\%$ CI is a paired bootstrap
  ($5{,}000$ resamples) and $p$ a two-sided exact McNemar test on the discordant
  pairs. The paradox ($\STI{}\rightarrow\SIT{}$) is significant everywhere except
  on Gemma-3.}
  \label{tab:significance}
  \centering
  \small
  \begin{tabular}{@{}lccc@{}}
    \toprule
    Comparison & $\Delta$ & 95\% CI & McNemar $p$\\
    \midrule
    \multicolumn{4}{@{}l}{\emph{Question-first paradox} ($\STI{}\rightarrow\SIT{}$)}\\
    \multicolumn{4}{@{}l}{\quad\emph{NaturalBench} (group)}\\
    \quad Qwen3-VL-8B & $+0.081$ & $[+0.058,+0.103]$ & $<$$10^{-4}$\\
    \quad Qwen2.5-VL-7B & $+0.176$ & $[+0.155,+0.196]$ & $<$$10^{-4}$\\
    \quad InternVL3-8B & $+0.079$ & $[+0.054,+0.105]$ & $<$$10^{-4}$\\
    \quad LLaVA-1.5-7B & $+0.123$ & $[+0.109,+0.138]$ & $<$$10^{-4}$\\
    \quad Gemma-3-27B & $-0.006$ & $[-0.026,+0.015]$ & $.620$\\
    \multicolumn{4}{@{}l}{\quad\emph{POPE} (per-question)}\\
    \quad Qwen3-VL-8B & $+0.020$ & $[+0.015,+0.025]$ & $<$$10^{-4}$\\
    \quad Qwen2.5-VL-7B & $+0.056$ & $[+0.050,+0.063]$ & $<$$10^{-4}$\\
    \quad InternVL3-8B & $+0.053$ & $[+0.046,+0.060]$ & $<$$10^{-4}$\\
    \quad LLaVA-1.5-7B & $+0.371$ & $[+0.358,+0.385]$ & $<$$10^{-4}$\\
    \quad Gemma-3-27B & $+0.005$ & $[-0.001,+0.012]$ & $.124$\\
    \multicolumn{4}{@{}l}{\quad\emph{Winoground} (group)}\\
    \quad Qwen3-VL-8B & $+0.095$ & $[+0.050,+0.140]$ & $<$$10^{-4}$\\
    \quad Qwen2.5-VL-7B & $+0.185$ & $[+0.145,+0.225]$ & $<$$10^{-4}$\\
    \quad InternVL3-8B & $+0.210$ & $[+0.163,+0.258]$ & $<$$10^{-4}$\\
    \quad LLaVA-1.5-7B & $+0.048$ & $[+0.028,+0.070]$ & $<$$10^{-4}$\\
    \quad Gemma-3-27B & $-0.005$ & $[-0.052,+0.043]$ & $.916$\\
    \midrule
    \multicolumn{4}{@{}l}{\emph{Echoing ladder} (NaturalBench group)}\\
    \quad Qwen3-VL-8B: $\SIT{}\rightarrow\STIT{}$ & $-0.001$ & $[-0.022,+0.020]$ & $1.000$\\
    \quad Qwen3-VL-8B: $\STIT{}\rightarrow\SITIT{}$ & $+0.024$ & $[+0.005,+0.043]$ & $.016$\\
    \quad Gemma-3-27B: $\SIT{}\rightarrow\STIT{}$ & $+0.026$ & $[+0.006,+0.046]$ & $.014$\\
    \quad Gemma-3-27B: $\STIT{}\rightarrow\SITIT{}$ & $+0.003$ & $[-0.015,+0.021]$ & $.819$\\
    \bottomrule
  \end{tabular}
\end{table}

\paragraph{NaturalBench sub-metrics.} \Cref{tab:naturalbench-full}
expands the NaturalBench column of the main results table
(\cref{tab:main}) into all four official metrics: Group (all four
(image, question) pairs in a group correct), Question (both images
correct for a question), Image (both questions correct for an image), and
Pair (a single (image, question) pair correct). The position ladder of
the main text holds on every sub-metric for every model.

\begin{table}[htb]
  \caption{NaturalBench (1{,}900 groups), all four metrics. The ordering
  ladder of the main text (\STI{} $<$ \SIT{} $\approx$ \STIT{} $<$
  \SITIT{}) holds on every sub-metric, except on Qwen2.5-VL, where question
  echoing alone under-recovers and the image copy is needed to match
  question-last. Best per model in \textbf{bold}.}
  \label{tab:naturalbench-full}
  \centering
  \small
  \begin{tabular}{@{}llcccc@{}}
    \toprule
    Model & Ordering & Group & Question & Image & Pair\\
    \midrule
    \multirow{5}{*}{Qwen3-VL-8B}
      & \STI{} (Q-first) & 0.270 & 0.523 & 0.560 & 0.753\\
      & \SIT{} (Q-last)  & 0.351 & 0.600 & 0.621 & 0.792\\
      & \STIT{} (ours) & 0.350 & 0.593 & 0.622 & 0.790\\
      & \SITIT{} (ours)         & \textbf{0.374} & \textbf{0.619} & \textbf{0.643} & \textbf{0.804}\\
      & \SITITr{} (ours)        & \textbf{0.374} & 0.617 & \textbf{0.643} & 0.803\\
    \midrule
    \multirow{5}{*}{Gemma-3-27B}
      & \STI{} (Q-first) & 0.232 & 0.490 & 0.532 & 0.732\\
      & \SIT{} (Q-last)  & 0.226 & 0.491 & 0.526 & 0.733\\
      & \STIT{} (ours) & 0.253 & \textbf{0.516} & \textbf{0.551} & \textbf{0.746}\\
      & \SITIT{} (ours)         & \textbf{0.255} & 0.513 & 0.546 & \textbf{0.746}\\
      & \SITITr{} (ours)        & 0.252 & 0.506 & 0.540 & 0.740\\
    \midrule
    \multirow{4}{*}{Qwen2.5-VL-7B}
      & \STI{} (Q-first) & 0.102 & 0.313 & 0.336 & 0.649\\
      & \SIT{} (Q-last)  & \textbf{0.277} & \textbf{0.542} & \textbf{0.562} & \textbf{0.760}\\
      & \STIT{} (ours)   & 0.191 & 0.451 & 0.477 & 0.718\\
      & \SITIT{} (ours)  & 0.276 & 0.532 & 0.551 & 0.757\\
    \midrule
    \multirow{4}{*}{InternVL3-8B}
      & \STI{} (Q-first) & 0.276 & 0.530 & 0.568 & 0.756\\
      & \SIT{} (Q-last)  & 0.355 & 0.605 & 0.627 & 0.796\\
      & \STIT{} (ours)   & 0.365 & 0.611 & 0.633 & 0.798\\
      & \SITIT{} (ours)  & \textbf{0.393} & \textbf{0.631} & \textbf{0.649} & \textbf{0.808}\\
    \midrule
    \multirow{3}{*}{LLaVA-1.5-7B}
      & \STI{} (Q-first) & 0.000 & 0.017 & 0.035 & 0.503\\
      & \SIT{} (Q-last)  & \textbf{0.123} & \textbf{0.365} & 0.413 & \textbf{0.669}\\
      & \STIT{} (ours)   & 0.120 & 0.357 & \textbf{0.420} & 0.663\\
    \bottomrule
  \end{tabular}
\end{table}

\paragraph{Random-permutation control (\SITITp{}).} Replacing the second
image copy's reversal with a random permutation of its patches (2D
positional indices preserved) matches both forward and reversed
re-presentation on all three benchmarks and both models: NaturalBench
group 0.380 / 0.249, POPE accuracy 0.888 / 0.834, and Winoground group
0.403 / 0.312 (Qwen3-VL / Gemma-3), each within noise of \SITIT{}. The
gain of re-presentation is the second whole-image read itself, not any
particular order of the second copy.

\section{Supplementary: the reversal follows the mask (\SITITr{})}
\label{sec:reversal}
Does the second image copy need to be \emph{reversed}? The mechanism says
no, and it says so differently for each model, which doubles as a check.
Forward repetition already supplies whole-image context, so reversing the
second copy's patch order and positional indices (\SITITr{}) should add
only scan-order diversity: a small gain for a decoder that is causal over
image tokens, nothing for one that is already bidirectional. Reading the
attention masks from each model's code, Qwen3-VL is causal over image
tokens and Gemma-3 is bidirectional within each image block
(\cref{fig:attn}). The prediction holds: reversal helps Qwen3-VL slightly
(Winoground group 0.403 to 0.410) and slightly hurts Gemma-3 (0.318 to
0.298). The probes locate the reversal effect in perception, not
read-out: \SITITr{} matches \SITIT{} on answer-to-question attention and
answer emergence (final P(correct) 0.85 vs.\ 0.84). A final control
replaces the reversal with a \emph{random}
permutation of the second copy with 2D positions preserved (\SITITp{}):
it matches forward and reversed re-presentation on all three benchmarks
and both models (\cref{sec:supp-results}), so the gain is the second whole-image
read itself, not any particular order of the second copy.

\begin{finding}
\textbf{Finding.} The mechanism predicts the sign of the reversal effect
per model, and the prediction holds: reversal helps the causal Qwen3-VL
decoder slightly (Winoground 0.403 to 0.410) and not the
already-bidirectional Gemma-3 (0.318 to 0.298); the sign tracks the mask
read from each model's code. A random second-copy order (\SITITp{})
matches both, so the gain is the second whole-image read, not any
particular order of it.
\end{finding}

\section{Supplementary: what the attention knockout does and does not remove}
\label{sec:supp-knockout}
The causal intervention of \cref{sec:causal} deserves one clarification, because
a natural reading is that forbidding the answer to attend to the question should
leave the model no way to answer at all. It does not, for two reasons.

\paragraph{The question stays in the prompt.} We do not delete the question or
mask it globally. We add a pre-softmax $-\infty$ only to the entries of the
attention matrix that connect the \emph{answer position} (the final token, in the
reported ``last'' scope) to the \emph{question-token columns}, at every text
layer. Every other position still attends to the question freely; the question is
still embedded and still shapes the rest of the computation. The answer-suffix
(``\dots\ Answer Yes or No:'') is untouched, so the model always emits a
well-formed yes/no token. The knockout changes \emph{which} answer the model
gives, not \emph{whether} it answers.

\paragraph{The question's content still reaches the answer, indirectly.} In a
transformer a token's meaning does not live only in its own column. Over the
early layers, attention copies the question's content forward into the hidden
states of the tokens that sit between the question and the answer (the trailing
punctuation and the answer-suffix tokens). Those relay tokens attend to the
question freely, since the knockout touches only the answer position's edges. The
answer token can then attend to the relay tokens, so the question's content
reaches it through a two-hop path (question $\rightarrow$ relay token
$\rightarrow$ answer) and through the residual stream, even though its direct edge
is cut. This forward movement of information is a well-documented property of
transformer computation.

\paragraph{What the number therefore measures.} The knockout isolates the
\emph{marginal} causal contribution of the answer's \emph{direct} read of the
question, on top of whatever leaks through relays; it is not a measure of total
question access. This is why severing an edge the answer does use lowers accuracy
only partially (question-last $0.588\rightarrow0.532$, toward chance on the
question-dependent items) rather than to random output, and it is why the
reportable claim is the
\emph{dissociation} (the direct edge is causal under question-last and inert under
question-first, where the answer instead reads the image), not a full-collapse
claim. The same relay effect is why the broader ``downstream'' knockout, which
also severs the relay tokens, behaves less cleanly and is reported only as
corroboration.

\section{Supplementary: the perception probe, mapped}
\label{sec:supp-cosine}

\begin{figure}[htb]
  \centering
  \includegraphics[height=4.4cm]{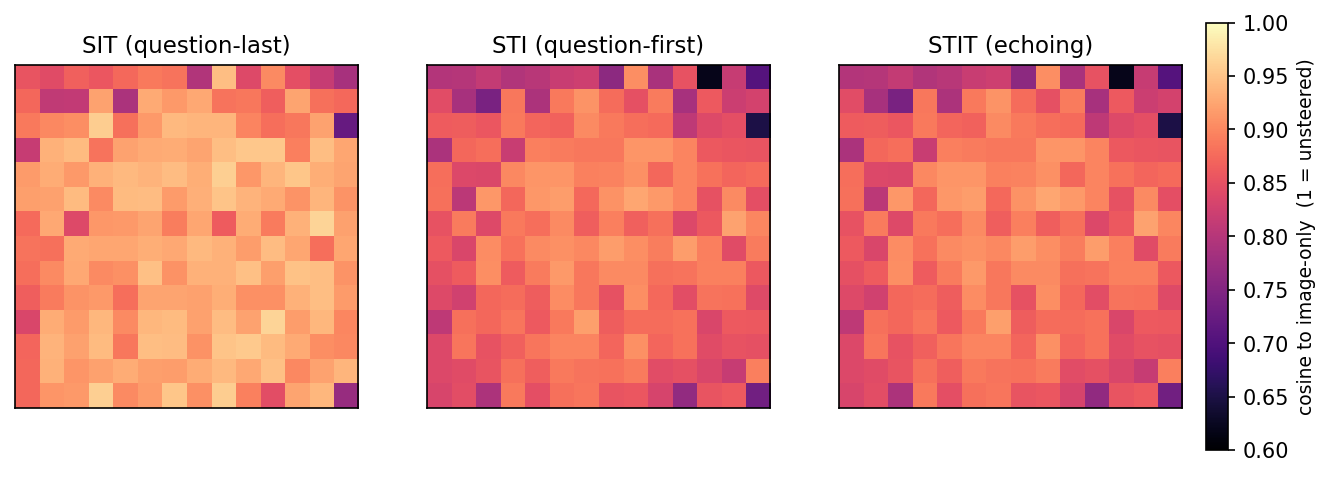}
  \caption{\textbf{Steering, measured, across orderings.} Per-patch
  cosine between image-token hidden states (layer 18) and the image shown
  alone, averaged over NaturalBench images (darker = moved further).
  Question-last (\SIT{}) barely steers the image (mean 0.91);
  question-first (\STI{}) steers it markedly (mean 0.86, down to 0.62);
  echoing (\STIT{}) is identical to \STI{} (cosine 1.000) because the
  causal mask hides the post-image question from the patches. Steering
  comes only from a question that precedes the image, so echoing acts
  purely at read-out.}
  \label{fig:cosine}
\end{figure}

\Cref{fig:cosine} maps the per-patch steering measurement summarized in
\cref{sec:steering}: only a question that precedes the image moves the
patch encodings, and echoing reproduces question-first's steering
exactly.

\section{Supplementary: the gap scales with image-token count}
\label{sec:supp-scaling}

\begin{figure}[htb]
  \centering
  \includegraphics[width=\linewidth]{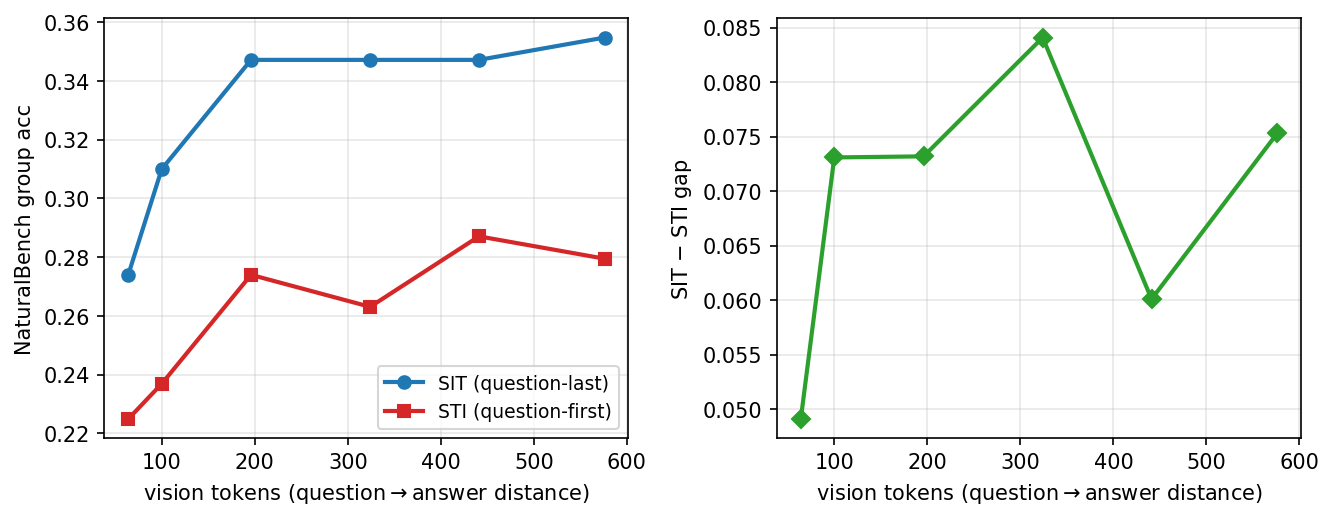}
  \caption{\textbf{The question-first gap scales with the number of image
  tokens.} NaturalBench group accuracy (Qwen3-VL, 916 images with short
  side $\geq$ 784\,px). \emph{Left:} question-last (\SIT{}) saturates
  near 0.35 once the image is legible; question-first (\STI{}) stays
  lower and keeps climbing. \emph{Right:} the \SIT{}$-$\STI{} gap widens
  with token count (0.05 at 64 tokens to 0.08 at 324), then plateaus as
  \SIT{} saturates. More image tokens push the question further from the
  answer under \STI{}, while \SIT{} keeps it adjacent.}
  \label{fig:tokencount}
\end{figure}

\Cref{fig:tokencount} gives the full resolution sweep behind
\cref{sec:scaling}. We re-render NaturalBench at a range of resolutions,
which sets the vision-token count, using only the 916 images whose short
side exceeds the largest rendered resolution (784\,px) so that every
point is a genuine downscale and no image is upscaled. Question-last
(\SIT{}) saturates near 0.35 group accuracy once the image is legible;
question-first (\STI{}) stays lower and keeps climbing as better
perception partly offsets its read-out penalty. The \SIT{}$-$\STI{} gap
widens with token count in the distance-dominated regime (0.05 at 64
tokens to 0.08 at 324), then plateaus as \SIT{} saturates, matching the
read-out account's prediction in direction and shape.

\section{Supplementary: is the question-first rewrite localized?}
\label{sec:supp-qq}

The main text shows that question-first prompting steers the image
representation (\cref{sec:analysis}): intermediate patch encodings become more
question-relevant. A natural question is whether this rewrite is
\emph{spatially selective}, that is, whether asking about one object rewrites
the patches on that object more than others. We test this directly on a single
image with two questions, and find that the rewrite is real but
\emph{diffuse}: it is a global, question-conditioned shift of the whole visual
field rather than a spotlight on the queried object. We show this on a single image, as an
illustration of \emph{how} the steering distributes rather than a claim about how
typical the pattern is. The diffuseness may also be architectural: Qwen3-VL
injects visual features into the decoder at multiple depths (its DeepStack
design~\cite{QwenVL3Card}), which spreads image information across layers and
could blur an object-localized rewrite into a global one. We read the result as
suggestive and do not build on it.

\paragraph{Setup.} We use one image (a living room with a television showing
football and a sleeping cat) and two questions, $q_1=$``What sport is on
TV?'' and $q_2=$``What is the cat doing?''. For an ordering we take the
image-token hidden states $h(q)$ at every layer and, per patch, measure
$\cos(h(q_1), h(q_2))$: how much does \emph{which question you ask} change each
visual patch. Under an image-first ordering the decoder mask prevents the image
tokens from ever attending to the (later) question, so the image representation
cannot depend on the question; under question-first (\STI{}) the question
precedes the image and can rewrite it.

\begin{figure}[htb]
  \centering
  \includegraphics[width=0.72\linewidth]{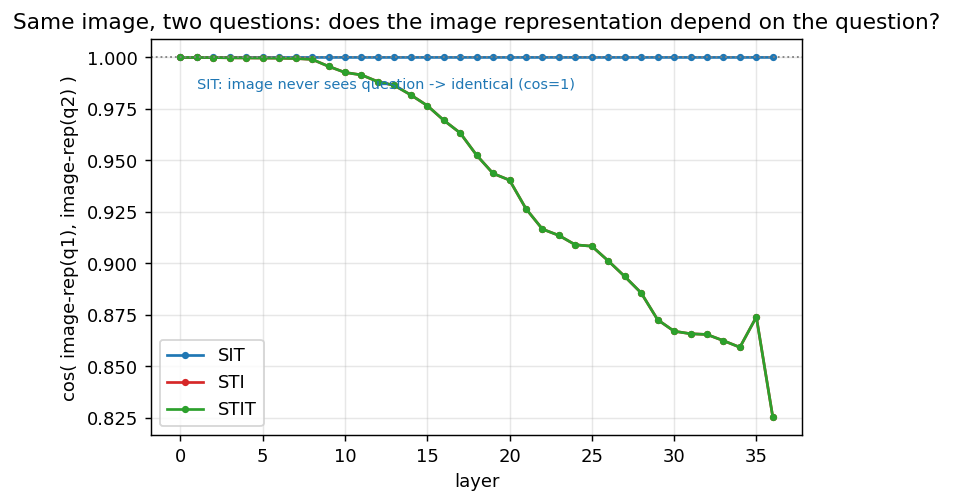}
  \caption{Same image, two questions. Per-layer $\cos$ between the image-token
  representations under $q_1$ and $q_2$. Image-first (blue) stays at $1.000$
  (the image never sees the question, so its encoding is question-invariant);
  question-first \STI{}/\STIT{} (red under green, identical curves) falls to
  $0.825$ (the question rewrites the image). This is a scalar, not a spatial,
  effect (see \cref{fig:qq-diffuse}).}
  \label{fig:qq-curve}
\end{figure}

\paragraph{The representation is question-dependent only under question-first.}
\Cref{fig:qq-curve} confirms the mask prediction exactly. Image-first keeps
$\cos(h(q_1),h(q_2))=1.000$ at every layer: the visual representation is
completely question-invariant. Question-first (\STI{}, and equivalently
question-echoing \STIT{}) drives the cosine down with depth to $0.825$ at the
final layer: the same image is encoded differently depending on what is asked.
This is the representation-level counterpart of the steering result in
\cref{sec:analysis}.

\begin{figure}[htb]
  \centering
  \includegraphics[width=\linewidth]{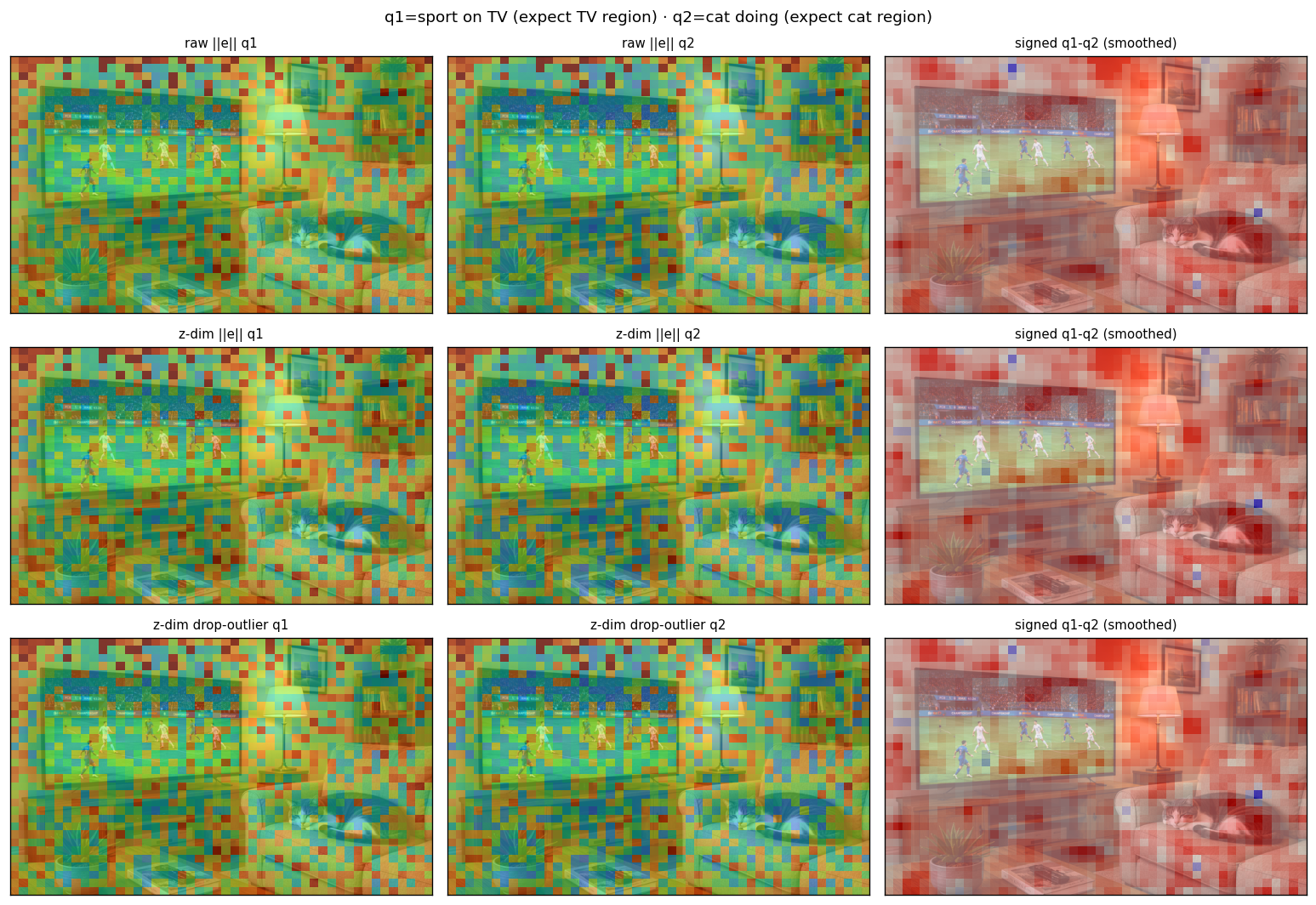}
  \caption{Per-patch question-first perturbation of the image representation
  (layer $18$). Rows: raw magnitude $\lVert e_q\rVert$ (top, dominated by
  outlier-dimension speckle), per-dimension standardized (middle), and
  standardized with the $20$ largest-variance dimensions dropped (bottom).
  Columns: $q_1$ magnitude, $q_2$ magnitude, and the signed difference
  $\lVert e_{q_1}\rVert-\lVert e_{q_2}\rVert$ (smoothed; red $=q_1$ rewrote more,
  blue $=q_2$). After de-noising, the $q_1$ and $q_2$ maps remain almost
  identical (correlation $0.98$) and the signed difference is a near-uniform
  field rather than a spotlight on the TV ($q_1$) or the cat ($q_2$): the rewrite
  is diffuse, not object-localized.}
  \label{fig:qq-diffuse}
\end{figure}

\paragraph{But the rewrite does not localize on the queried object.} We asked
whether the question-first rewrite concentrates on the relevant object (the TV
for $q_1$, the cat for $q_2$). It does not. Writing $e_q = h_{\STI}(q) - b$ for
the change each question makes relative to the image-only baseline $b$, two
facts emerge. First, the two questions' effect vectors are $92$--$96\%$ the same
vector: the shared, question-agnostic ``a question is present'' component
dominates, and only about $5\%$ of $e_q$ is specific to which question was
asked. Consequently the per-question perturbation maps $\lVert e_{q_1}\rVert$
and $\lVert e_{q_2}\rVert$ are nearly identical (spatial correlation $0.98$).
Second, the raw per-patch magnitude is dominated by a few outlier
``massive-activation'' hidden dimensions (per-patch coefficient of variation
$6.8$, salt-and-pepper speckle); standardizing each dimension removes this
(coefficient of variation $0.69$). Even after removing both the shared component
and the outlier dimensions, the residual question-specific perturbation is
spatially diffuse (\cref{fig:qq-diffuse}), not concentrated on the queried
object; a gradient-based (Grad-CAM) attribution of the answer token is likewise
dominated by background and border patches. The question-first steering is a
broad rewrite of the visual field, consistent with the main-text account in
which the paradox lies in downstream read-out rather than in mislocalized
perception. The scalar curve of \cref{fig:qq-curve}, not a per-patch heatmap, is
the faithful visualization of this effect.

\begin{figure}[htb]
  \centering
  \includegraphics[width=\linewidth]{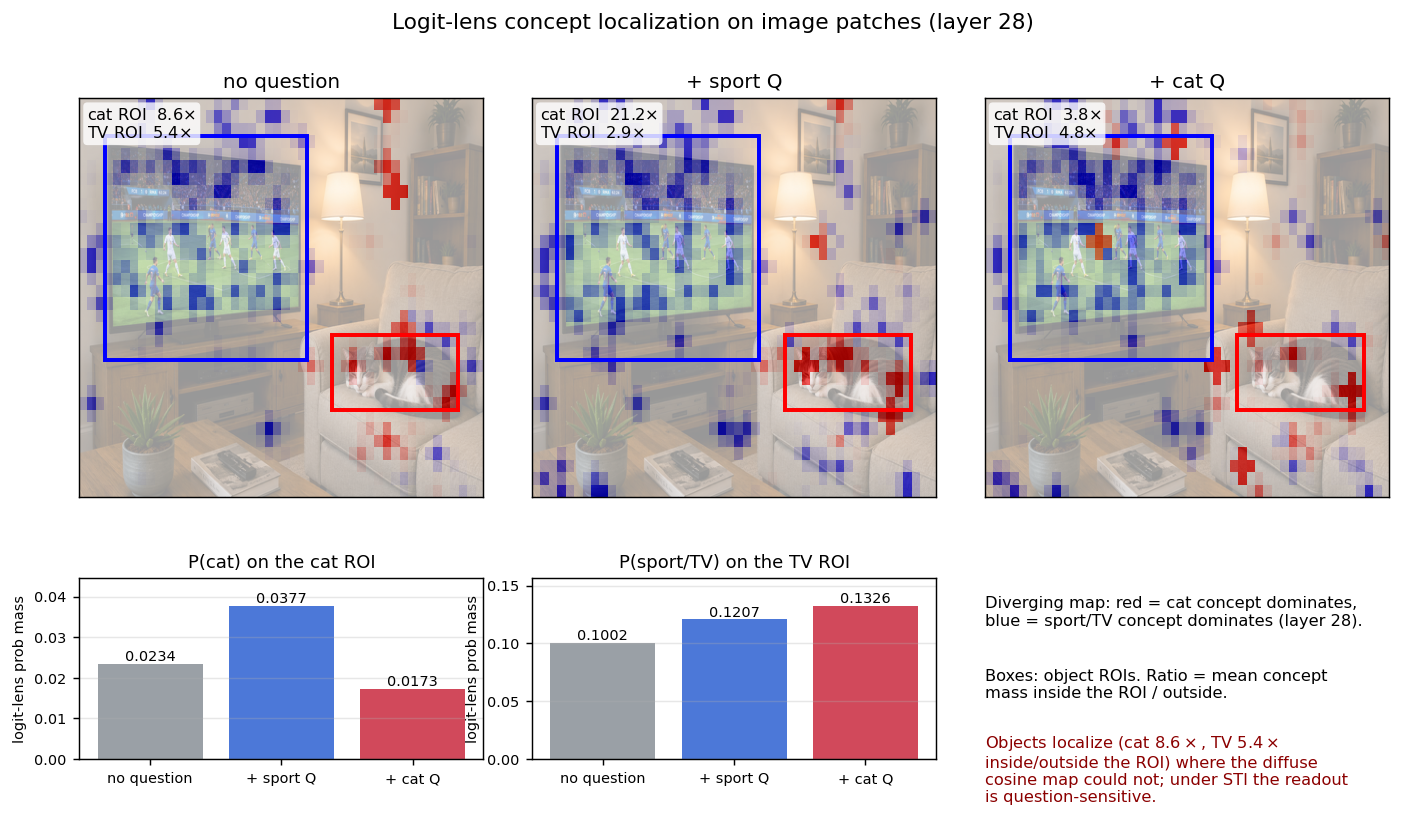}
  \caption{Per-patch logit-lens concept localization (layer $28$). Top: the
  diverging map $P(\text{cat})-P(\text{sport/TV})$ overlaid on the image (red: cat
  concept dominates; blue: sport/TV concept dominates), for the image-only
  baseline and question-first under each question; boxes are the object ROIs and
  the corner text is the inside/outside mass ratio. Bottom: raw logit-lens
  probability mass inside each object's ROI across conditions. Unlike the diffuse
  cosine map (\cref{fig:qq-diffuse}), the semantic readout localizes each object,
  and under question-first the readout on each object is question-sensitive.}
  \label{fig:qq-logitlens}
\end{figure}

\begin{figure}[htb]
  \centering
  \includegraphics[width=\linewidth]{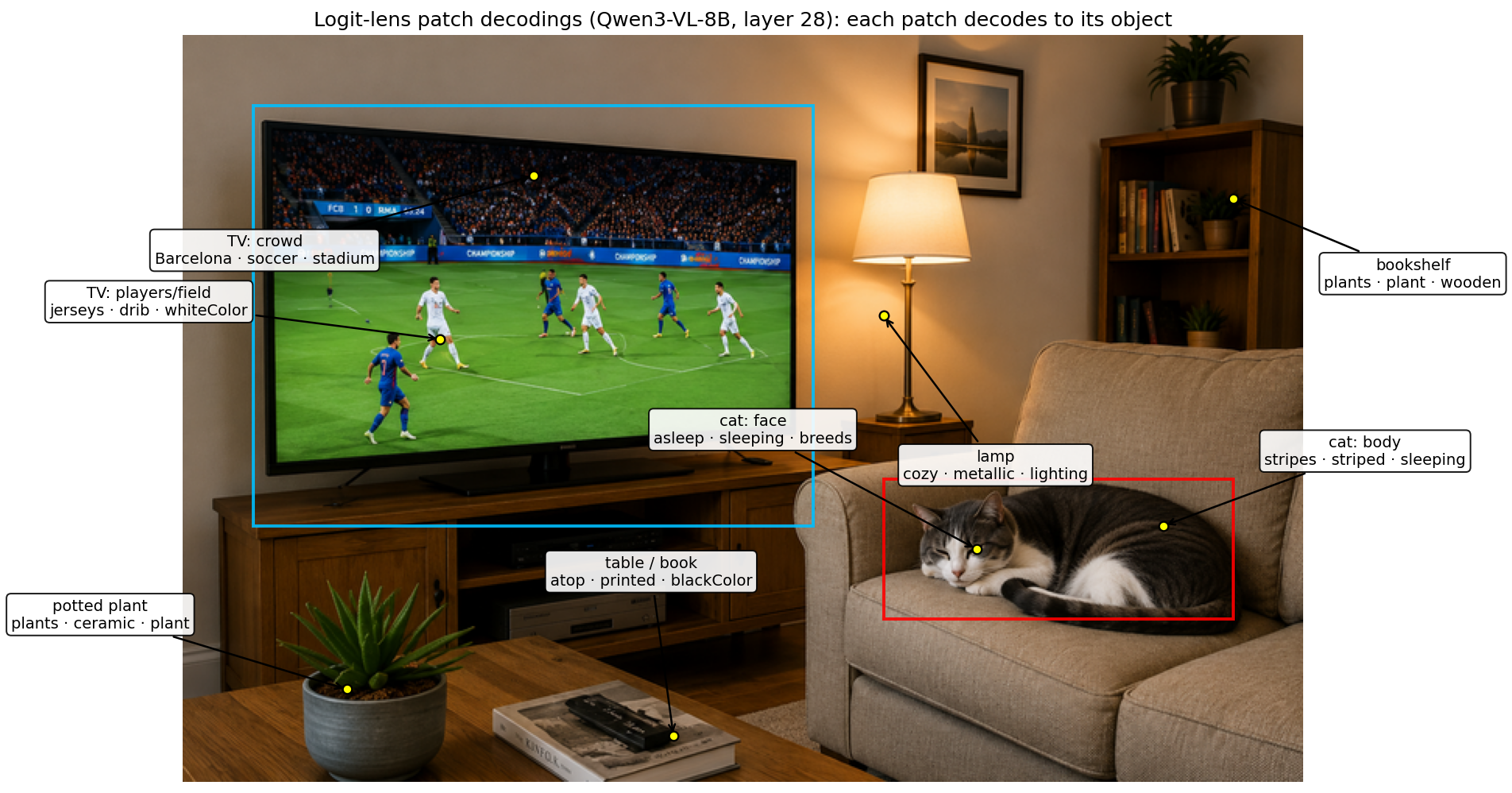}
  \caption{Logit-lens patch decodings (Qwen3-VL-8B, layer $28$). Each callout
  gives the top vocabulary tokens that a patch decodes to under the logit lens.
  The patches decode to their objects: the television crowd to
  \emph{Barcelona}/\emph{soccer}/\emph{stadium}, the players to \emph{jerseys},
  the lamp to \emph{cozy}/\emph{lighting}, the cat to \emph{asleep}/\emph{striped},
  and the potted plant to \emph{ceramic}/\emph{plant}. The visual representation
  carries localized, human-readable object identity, which is what the read-out
  stage must access; the question-first paradox is a failure to read this out, not
  to form it.}
  \label{fig:qq-logitlens-text}
\end{figure}

\begin{figure}[htb]
  \centering
  \includegraphics[width=\linewidth]{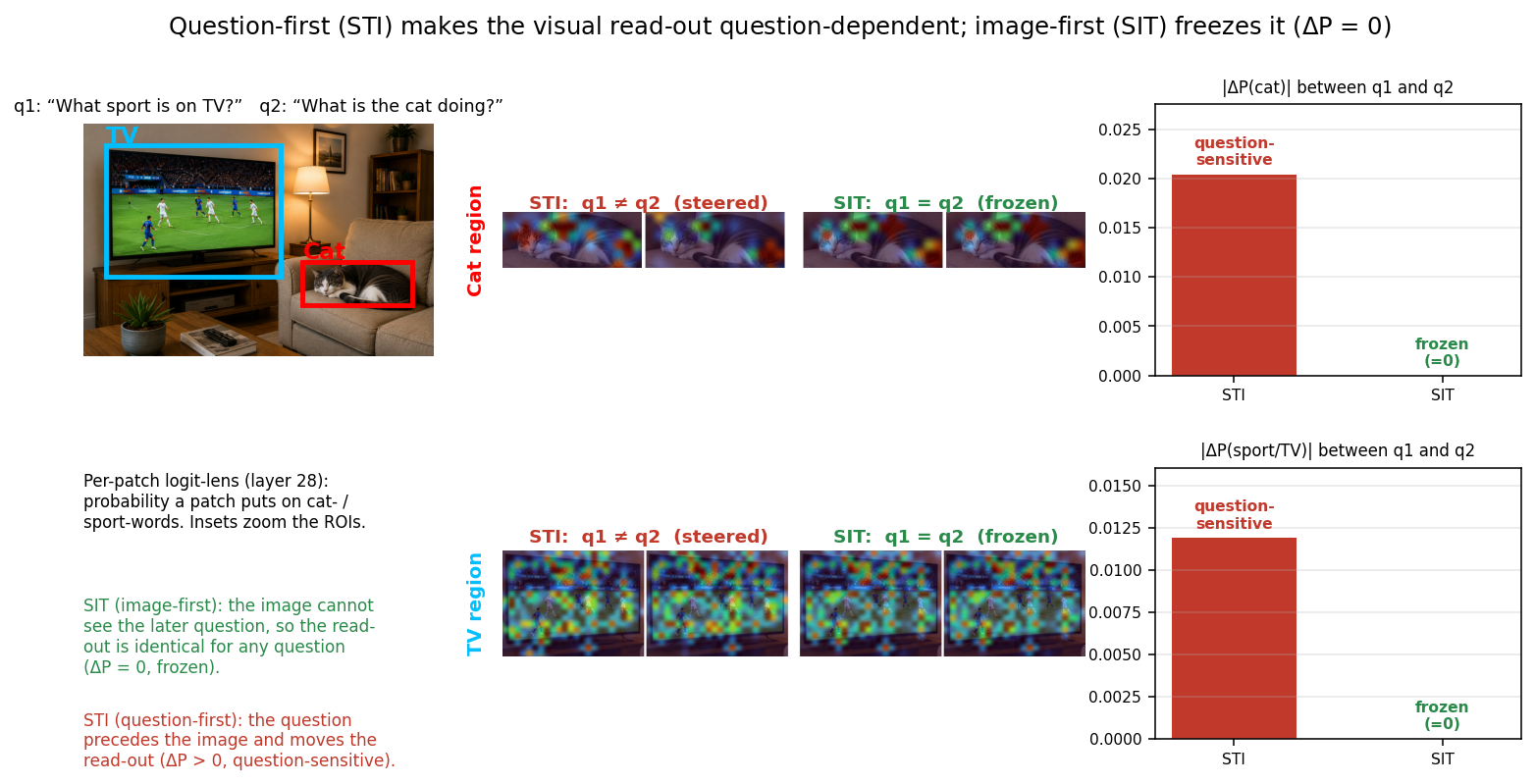}
  \caption{Question-first steers the visual read-out; image-first cannot. Same
  image, two questions ($q_1$ sport, $q_2$ cat). Insets zoom the per-patch
  logit-lens concept probability on the cat and TV ROIs; bars give the
  question-effect magnitude $|\Delta P|$ between the two questions on each ROI.
  Under \STI{} the read-out differs between questions ($|\Delta P|>0$,
  question-sensitive); under \SIT{} the image cannot attend to the later question,
  so the read-out is identical ($\Delta P = 0$, frozen). We show the magnitude of
  the question-effect, not its sign, which is not stable on one image. Single
  image, layer $28$; illustrative of the mechanism, not a dataset-level
  measurement.}
  \label{fig:qq-teaser}
\end{figure}

\paragraph{A semantic logit-lens readout does localize.} The diffuseness above
is a property of the raw representation geometry, not of the model's percept.
Projecting each image patch through the logit lens (final norm and unembedding)
and summing the probability on a concept's token set recovers a localized,
object-level readout. \Cref{fig:qq-logitlens} plots the per-patch probability of
cat-words minus sport/TV-words at layer $28$: the television reads as sport/TV
(blue) and the sleeping cat reads as cat (red), with each concept's mass
concentrated inside its object box (cat $8.6\times$, sport/TV $5.4\times$ more
mass inside the box than outside). Decoding each patch to its top vocabulary
token tells the same story in words (\cref{fig:qq-logitlens-text}): the crowd
patches read \emph{Barcelona}/\emph{soccer}/\emph{stadium}, the players
\emph{jerseys}, the cat \emph{asleep}/\emph{striped}, the potted plant
\emph{ceramic}/\emph{plant}. The readout is also question-sensitive under
question-first: the concept mass on each object ROI changes between the two
questions under \STI{}, whereas under image-first it is identical
(\cref{fig:qq-teaser}). The question-first rewrite is thus diffuse in the
residual stream yet semantically localized once projected to token space. This is
exactly what the read-out account of the main text predicts: perception localizes
objects correctly, so the paradox is a downstream failure to read the correctly
perceived answer, not a failure to perceive.

\paragraph{The read-out is question-dependent only under question-first.}
\Cref{fig:qq-teaser} makes the ordering contrast explicit on the same image.
Under image-first (\SIT{}) the image tokens precede the question and are masked
from it, so the per-patch logit-lens read-out is bit-identical for the two
questions ($\Delta P = 0$ exactly). Under question-first (\STI{}) the question
precedes the image and moves the read-out: the concept mass on each object ROI
differs between the two questions ($|\Delta P|>0$). We report the magnitude of
this question-effect rather than its sign, which is not stable on a single image;
the robust, mask-level fact is that question-first makes the visual read-out
question-dependent while image-first freezes it. This is the localized, semantic
counterpart of the scalar cosine result (\cref{fig:qq-curve}).

\section{Supplementary: system prompt placement (\IST{} vs \SIT{})}
\label{sec:supp-sysfirst}

Every ordering in the main text prepends the fixed system message. Here we
justify that default by moving only the system block: \IST{} (Image, System,
Task) leads with the image, while \SIT{} (System, Image, Task) leads with the
system prompt. The two are anagrams that keep the question in the same
question-last position, so any difference isolates the effect of \emph{where the
system prompt sits} relative to the image.

\paragraph{System-first helps.} On Qwen3-VL-8B, placing the system prompt before
the image (\SIT{}) beats image-first (\IST{}) on all three benchmarks
(\cref{tab:sysfirst}): NaturalBench group accuracy rises $+0.011$
($0.339\!\to\!0.350$), POPE accuracy $+0.002$, and Winoground group accuracy
$+0.038$ ($0.280\!\to\!0.318$). The gains are small but uniformly positive across
datasets and metrics, which is why we prepend the system message in every
ordering we report; it also matches the standard chat-template convention the
model is trained with.

\paragraph{Relation to the steering analysis.} This connects to \cref{fig:cosine}.
Under \IST{} the image is the very first content, so under the causal mask its
patch encodings are identical to the image-only baseline (cosine exactly
$1.000$). Under \SIT{} the system prompt precedes the image, so the image tokens
attend back to it and their encoding moves slightly off image-only (mean cosine
$0.91$). The accuracy numbers show that this system-conditioned shift is benign to
mildly helpful on Qwen: perturbing the image with a task-agnostic system prefix
does not cost accuracy, unlike the question-first rewrite, whose damage is a
read-out failure rather than a perceptual one (\cref{sec:analysis}).

\begin{table}[htb]
  \caption{System prompt placement on Qwen3-VL-8B, question held question-last.
  \SIT{} (system first) vs \IST{} (image first); $\Delta=\SIT{}-\IST{}$.
  System-first helps on every benchmark. NaturalBench and Winoground are group
  accuracy, POPE is accuracy.}
  \label{tab:sysfirst}
  \centering
  \small
  \begin{tabular}{@{}lccc@{}}
    \toprule
    Benchmark (metric) & \IST{} & \SIT{} & $\Delta=\SIT{}-\IST{}$\\
    \midrule
    NaturalBench (Group) & 0.339 & 0.350 & $\mathbf{+0.011}$\\
    POPE (Acc)           & 0.888 & 0.891 & $+0.002$\\
    Winoground (Group)   & 0.280 & 0.318 & $\mathbf{+0.038}$\\
    \bottomrule
  \end{tabular}
\end{table}

\begin{finding}
\textbf{Finding.} Prepending the system prompt (\SIT{}) rather than leading with
the image (\IST{}) gives a small, consistent accuracy gain on Qwen3-VL-8B across
all three benchmarks (up to $+0.038$ Winoground group accuracy), so we default to
system-first. The system prefix perturbs the image encoding only mildly (cosine
$0.91$ vs.\ $1.000$) and benignly, unlike the question-first rewrite.
\end{finding}

\end{document}